\documentclass[journal]{IEEEtran}
\usepackage{ifpdf}
\usepackage{cite}
\ifCLASSINFOpdf
\usepackage[pdftex]{graphicx}
\graphicspath{{../imgs/}}
\else
  \usepackage[dvips]{graphicx}
  \graphicspath{{../imgs/}}
\fi
\usepackage{amsmath}
\usepackage{algorithmic}
\usepackage{array}
\ifCLASSOPTIONcompsoc
  \usepackage[caption=false,font=normalsize,labelfont=sf,textfont=sf]{subfig}
\else
  \usepackage[caption=false,font=footnotesize]{subfig}
\fi
\usepackage{stfloats}
\usepackage{url}
\usepackage{commath}
\usepackage{amssymb}
\usepackage[]{footmisc}
\usepackage{multirow}
\usepackage{hyperref}
\usepackage[caption=false]{subfig}

\begin{document}

\title{Hyperspectral Pansharpening Based on Improved Deep Image Prior and  Residual Reconstruction}

\author{Wele~Gedara~Chaminda~Bandara,~\IEEEmembership{Student Member,~IEEE,}
        Jeya~Maria~Jose~Valanarasu,~\IEEEmembership{Student Member,~IEEE,}
        and~Vishal~M.~Patel,~\IEEEmembership{Senior~Member,~IEEE}
\thanks{Wele Gedara Chaminda Bandara, Jeya Maria Jose Valanarasu and, Vishal M. Patel are with Whiting School of Engineering, The Johns Hopkins University, 3400 North Charles Street, Baltimore, MD 21218-2608 USA (email: wbandar1@jhu.edu; jvalana1@jhu.edu; vpatel36@jhu.edu).}
}

\markboth{IEEE TRANSACTIONS ON GEOSCIENCE AND REMOTE SENSING, VOL. xx, NO. xx, Month Year}%
{}



\maketitle

\begin{abstract}
    \par Hyperspectral pansharpening aims to synthesize a low-resolution hyperspectral image (LR-HSI) with a registered panchromatic image (PAN) to generate an enhanced HSI with high spectral and spatial resolution.  Recently proposed HS pansharpening methods have obtained remarkable results using deep convolutional networks (ConvNets), which typically consist of three steps: (1) up-sampling the LR-HSI, (2) predicting the residual image via a ConvNet, and (3) obtaining the final fused HSI by adding the outputs from first and second steps.  Recent methods have leveraged Deep Image Prior (DIP) to up-sample the LR-HSI due to its excellent ability to preserve both spatial and spectral information, without learning from  large data sets. However, we observed that the quality of up-sampled HSIs can be further improved by introducing an additional spatial-domain constraint to the conventional spectral-domain energy function. We define our spatial-domain constraint as the $L_1$ distance between the predicted PAN image and the actual PAN image. To estimate the PAN image of the up-sampled HSI, we also propose a learnable spectral response function (SRF). Moreover, we noticed that the residual image between the up-sampled HSI and the reference HSI mainly consists of edge information and very fine structures. In order to accurately estimate fine information, we propose a novel over-complete network, called HyperKite, which focuses on learning high-level features by constraining the receptive from increasing in the deep layers. We perform experiments on three HSI datasets to demonstrate the superiority of our DIP-HyperKite over the state-of-the-art pansharpening methods. The deployment codes, pre-trained models, and final fusion outputs of our DIP-HyperKite and the methods used for the comparisons will be publicly made available at \url{https://github.com/wgcban/DIP-HyperKite.git}
\end{abstract}


\ifCLASSOPTIONpeerreview
 \begin{center} \bfseries EDICS Category: 3-BBND \end{center}
\fi
\IEEEpeerreviewmaketitle

\section{Introduction}
\IEEEPARstart{H}{yperspectral} images (HSIs) with a large number of spectral bands have gained immense attention in the field of remote sensing due to its applications in broad research areas such as classification~\cite{hyperspctral_classification}, unmixing~\cite{hyper_unmixing}, anomaly detection~\cite{hyper_anomaly}, change detection~\cite{hyper_change_detection}, etc. However, due to the limited incident energy available when capturing an image, hyperspectral imaging systems face trade-offs between spectral resolution, spatial resolution, and signal-to-noise ratio (SNR)~\cite{hyperspectral_pan_review}. For this reason, hyperspectral imaging systems can provide images with high spectral resolution but with low spatial resolution. In contrast, multispectral imaging systems can provide data with high spatial resolution but with fewer spectral bands (e.g., panchromatic images or multispectral images(MSIs) with three or four spectral bands). Low spatial resolution in HSIs leads to relatively poor performance in some practical remote sensing applications, such as road topology extraction~\cite{road_from_panHSI}, and spectral unmixing~\cite{spectral_unmixing_pan}. Therefore, full-resolution HSIs with high spatial and spectral resolution are desired. One way to obtain such ideal HSIs is to fuse high spectral resolution HSIs with high spatial resolution PAN/MSIs. This fusion process is called HS pansharpening in the remote sensing literature, which is indeed a form of super-resolution.

\par Traditional pansharpening methods can be mainly divided into four classes~\cite{hyperspectral_pan_review}: (1) component substitution (CS), (2) multi-resolution analysis (MRA), (3) Bayesian, and (4) matrix factorization. Component substitution methods rely on substituting the spatial component of the HSI with the MSI/PAN image. The family of CS contains algorithms such as Gram–Schmidt adaptive (GSA)~\cite{GSA, GS}, principal component analysis (PCA)~\cite{PCA_1, PCA1, PCA2}, and intensity-hue-saturation (IHS)~\cite{IHS_1}. Even though the CS methods usually generate pansharpened HSIs with accurate spatial information, sometimes they suffer from critical spectral distortions. The MRA approaches are based on injecting the spatial details obtained through the multi-scale decomposition of the MSI/PAN image into the HSI. In order to extract the spatial details from the PAN image, several algorithms have been proposed in the literature, such as decimated wavelet transform (DWT)~\cite{DWT}, undecimated wavelet transform (UDWT)~\cite{UDWT}, smoothing filter-based intensity modulation (SFIM)~\cite{SFIM}, modulation transfer function with generalized Laplacian pyramid (MTF-GLP)~\cite{MTF-GLP}, and MTF-GLP with high-pass modulation (MTF-GLP-HPM)~\cite{MTF-GLP-HPM}. In contrast to the CS methods, the MRA family performs better in spectral preservation, but is more sensitive to registration errors which may cause critical distortions in the spatial domain. Due to these inherent advantages and disadvantages of CS and MRA approaches, there have been works which attempted to  combine both CS and MRA methods. One of the representatives of hybrid CS and MRA algorithm is guided filter PCA (GFPCA)~\cite{GFPCA}. The Bayesian-based methods also provide a convenient way to regularize the fusion methods by modeling the posterior distribution of the target HSI provided that the LR-HSI and MSI/PAN image. Examples of the algorithms based on the Bayesian inference framework include convex regularization under a Bayesian framework (abbreviated as Hysure) ~\cite{hysure}, naive Bayesian Gaussian prior (abbreviated as BF)~\cite{BF}, and sparsity promoted Gaussian prior (abbreviated as BFS)~\cite{BFS}. Finally, the coupled non-negative matrix factorization (abbreviated as CNMF) is one of the examples for matrix factorization-based methods, which regularizes the fusion problem by using the priors of spectral unmixing~\cite{CNMF}. However, the fusion performance of traditional pansharpening approaches is generally limited due to their inadequate representation ability. In addition, the algorithms mentioned above may result in severe quality degradation when the assumptions do not align with a particular dataset.  Furthermore, most traditional pansharpening approaches typically reach the optimal solution through an iterative process, which is time-consuming and inefficient. 

\par Recently, deep learning (DL) models based on convolutional neural networks (ConvNets) have also been introduced for the HS pansharpening problem due to ConvNets' excellent ability to learn high-level features automatically.  ConvNet-based HS pansharpening methods generally consist of three steps, 
\begin{enumerate}
    \item \textit{Up-sampling step}: Up-sampling the LR-HSI to the spatial resolution of the PAN image, 
    \item \textit{Residual reconstruction step}: Concatenating the up-sampled HSI and PAN image along the spectral dimension and passing it through a residual learning network to learn the residual image,
    \item \textit{Final fusion step}: Obtaining the final fused HSI by adding the up-sampled HSI and the residual image.
\end{enumerate}

\par There have been many  methods proposed to up-sample LR-HSI to the spatial resolution of PAN. In the earliest studies, nearest-neighbor and bicubic interpolation were the famous methods to perform up-sampling. However, the methods mentioned above conduct upsampling on each band of the LR-HSI successively, thus ignoring the high spectral correlation of HSIs which may lead to spectral distortions \cite{DHP-DARN, DeepImagePrior}. In order to minimize the spectral distortion, data-driven up-sampling techniques (i.e., deep super-resolution networks) have also been utilized in HS pansharpening. The LapSRN~\cite{LapSRN} network is an example of such a data-driven super-resolution method, which progressively super-resolves a LR image in a coarse-to-fine manner in a Laplacian pyramid framework. However, the LapSRN method requires a large number of images for training which is impractical in the HS domain due to the limited number of datasets available to the public.  A remedy to the problem mentioned above was proposed by Ulyanov \textit{et al.} \cite{DeepImagePrior} where they proposed a deep learning-based super-resolution framework called deep image prior (DIP). The proposed method uses a randomly initialized ConvNet to upsample an image, using its structure as an image prior, similar to bicubic upsampling. However, this method does not require any training but produces much cleaner results with sharper edges. Motivated by the super-resolution performance of DIP in the RGB domain, researchers have applied DIP to the HS pansharpening problem \cite{DeepHyperspectralPrior, DHP-DARN} and achieved impressive results. However, we observed that the energy function defined in HS DIP up-sampling directly applies the energy function formulated for the RGB DIP process, where they only impose spectral-domain constraint by computing the $L_1$ distance between the down-sampled version of the target up-sampled HSI and the LR-HSI. However, the existing HS DIP methods do not impose any spatial-domain constraint by utilizing the available PAN image. We address this issue by introducing an additional spatial-domain constraint to the HS DIP process as our first contribution.

\par For residual reconstruction, various ConvNet architectures have been proposed in the literature to accurately predict the residual component between the up-sampled HSI and the reference HSI with less spectral and spatial distortion. Among those, Giuseppe \textit{et al.} \cite{pansharpening_by_CNN} was the first to introduce simple three-layer ConvNet architecture for the residual learning. Further, Lin \textit{et al.} \cite{Hyper-PNN} improved the spatial and spectral prediction capability of Giuseppe's work (abbreviated as HyperPNN) by introducing spectral and spatial prediction modules. To further enhance the representational power of ConvNets, attention mechanisms \cite{attention_hyperspectral} have also been introduced. Among those, Zheng \textit{et al.} \cite{DHP-DARN} proposed a spatial and spectral attention mechanism (abbreviated as DHP-DARN) for the residual learning in which they cascade several channel-spatial-attention residual blocks to adaptively learn more informative channel-wise and spatial-domain features simultaneously. More recently, Xu \textit{et al.} \cite{SPDNet} proposed a design (abbreviated as SDPNet) based on two encoder-decoder networks to extract deep-level features from two types of source images with densely connected blocks
to strengthen feature propagation. However, we experimentally observed that most of the existing residual learning methods fail when predicting the high-frequency information, such as edges and delicate structures in the residual image. The main reason for this observation is due to the fact that the increasing receptive field of the network in the deep layers. Motivated by this observation, we introduce an over-complete network, called HyperKite, for residual reconstruction task as our second contribution, which constrains the receptive field from increasing in deep layers thus extracting more high-frequency information.

The main contributions of this paper are summarized as follows:
\begin{enumerate}
    \item A novel spatial constraint is introduced for the DIP up-sampling process. To the best of our knowledge, this is the first study that integrates both spatial and spectral constraints to the DIP up-sampling. The proposed spatial constraint significantly improves the spatial and spectral performance measures of the up-sampled HSIs.  
    \item An over-complete network, called HyperKite, is proposed for the residual reconstruction, which is highly capable of extracting high-frequency information of the residual image by appropriately constraining the receptive field of the network.
    \item We conduct extensive experiments to clearly demonstrate the improvements brought in from our contributions to the HS pansharpening. We compared the fusion performance of DIP-HyperKite with both conventional and deep learning-based approaches. The deployment codes, pre-trained models, and final fusion results of our DIP-HyperKite as well as the comparison methods in the results and discussion will be publicly made available at \url{https://github.com/wgcban/DIP-HyperKite.git}.
\end{enumerate}
The rest of this paper is organized as follows. Section \ref{sec: related_work} provides some basics DIP and over-complete representations. In Section \ref{sec: method} the proposed DIP-HyperKite is described in detail. Section \ref{sec: experimental_settings} describes the datasets and performance metrics that we used in the experiments. In Section \ref{sec: results}, the experimental results and discussions of different data sets are presented. Finally, the conclusions are drawn in Section \ref{sec: conclusion}.

\section{Related Work}
\label{sec: related_work}
\subsection{DIP for HSI up-sampling}
    \label{rw: dip}
    
    \par Generally, ConvNets have an excellent ability to learn realistic image priors from a large amount of visual data, placing them in leading positions on the benchmarks of various image processing tasks \cite{image_denoising, super_resolution}. Contrary to the general opinion on deep networks that they require large data to capture image priors, DIP \cite{DeepImagePrior} has shown that a randomly initialized network can capture low-level image statistics before any training. Concretely, in HS pansharpening, DIP can generate the up-sampled HSI $\mathbf{x}_{\text{dip}}$ of the LR-HSI $\mathbf{y}$ with spatial up-sampling factor $\beta$ by taking a fixed randomly initialized vector $\mathbf{z}$ as the input, and utilizing the deep network as a parametric function $\mathbf{x}_{\text{dip}} = f_{\theta} (\mathbf{z})$. Next, the network is optimized over its parameters $\theta$ to obtain the up-sampled HSI $\mathbf{x}_{\text{dip}}$ as follows:
    \begin{equation}
        \mathbf{x}_{\text{dip}} = \min_{\mathbf{x}_{\text{dip}}} Q(\mathbf{x}_{\text{dip}}; \mathbf{y}) + R(\mathbf{x}_{\text{dip}}),
        \label{eqn: SR_loss}
    \end{equation}
    where $Q(\mathbf{x}_{\text{dip}};\mathbf{y})$ is an energy function that controls the fidelity toward the LR-HSI $\mathbf{y}$, and $R(\mathbf{x}_{\text{dip}})$ is a regularization function based on prior knowledge. In \cite{DeepImagePrior}, it has been shown that the regularization term $R(\mathbf{x}_{\text{dip}})$ can be implicitly substituted by the deep network. Therefore, the minimization problem in (\ref{eqn: SR_loss}) has simplified to optimizing the network over its parameters $\theta$ as follows:
    \begin{equation}
        \theta^{*} = \arg \min_{\theta} Q(\mathbf{x}_{\text{dip}}; \mathbf{y}) \text{ s.t. } \mathbf{x}_{\text{dip}} = f_{\theta} (\mathbf{z}),
        \label{eq: simplified_q}
    \end{equation}
    where $\theta^{*}$denotes the optimal set of parameters of the network. Furthermore, the most straightforward and commonly utilized energy function in HS pansharpening is that the $L_1$ distance \cite{Hyper-PNN} between the down-sampled version of the up-sampled HSI $\mathbf{x}_{\text{dip}}$ and the LR-HSI $\mathbf{y}$ as follows:
    \begin{equation}
        Q(\mathbf{x}_{\text{dip}}; \mathbf{y}) = \norm{d(\mathbf{x}_{\text{dip}})-\mathbf{y}}_1 \text{ s.t. } \mathbf{x}_{\text{dip}} = f_{\theta}(\mathbf{z}),
        \label{eq: standard_loss}
    \end{equation}
    where $d(\cdot)$ denotes the down-sampling operator by a factor of $\beta$. 
    
\subsection{Over-complete ConvNets}
\begin{figure}[tb]
    \centering
    \includegraphics[width=\linewidth]{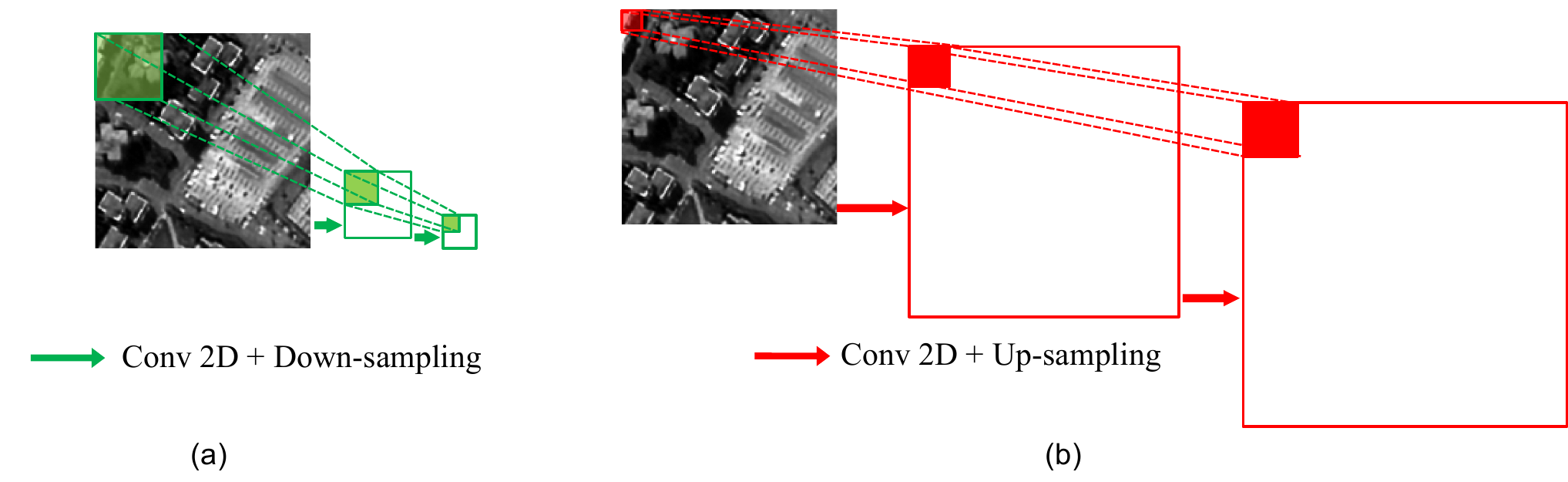}
    \caption{(a) Effect of under-complete ConvNet on receptive field where the deeper layers focus on a larger region of the input thus extracting high-level/low-frequency information. (b) Effect of over-complete ConvNet on receptive field where the deeper layers focus on a much smaller region in the input thus extracting low level/high-frequency information.}
    \label{fig:receptive_field}
\end{figure}
 Most of the current architectures in deep learning are  ``encoder-decoder'' \cite{ronneberger2015u, badrinarayanan2017segnet, he2016deep} based. Here, the encoder  translates the high-dimensional input to a low-dimensional latent space  while the decoder learns to take the latent low-dimensional representation back to a high-dimensional output.  These type of architectures learn low-level features at their initial layers and high-level features at their deeper layers. These are termed under-complete networks as the input is taken to a lower spatial dimension in the latent space.
 
 In signal processing, over-complete dictionaries are widely used for their highly robust characteristic \cite{lewicki2000learning}. The number of basis functions  here are more than the number of input signal samples which enables a higher flexibility for capturing structure in data. In \cite{vincent2008extracting}, over-complete auto-encoders were  found to be better feature extractors for denoising when compared to under-complete auto-encoders. In an over-complete network \cite{valanarasu2020kiu}, the encoder takes the input data to a higher spatial dimension unlike a traditional encoder. This is achieved by using an upsampling layer after every convolutional layer in the encoder. Using upsampling layers in the encoder causes the receptive field to be constrained in the deep layers. This causes the deep layers in the network to learn more fine-context high-frequency information when compared to under-complete networks. Increase in receptive field for an over-complete network can be generalized in an $i^{th}$ layer  as follows:  
 \begin{equation}
    RF (\text{w.r.t } I) =  \left(\frac{1}{2}\right)^{2(i-1)} \times k \times k,
 \end{equation}
 where the initial receptive field of the conv filter is assumed to be $k \times k$ on the image $I$. This phenomenon has been visualized in Fig \ref{fig:receptive_field}. As shown in Figure \ref{fig:receptive_field} (b), by employing an upsampling layer after every convolutional layer in the encoder, the over-complete network restricts the receptive field size to a smaller region which forces the network to learn very fine edges as it tries to focus heavily on smaller regions. This is completely different from the conventional over-complete architectures where they perform downsampling after each convolution block which makes the network to focus on a much larger region in the input as shown in Figure \ref{fig:receptive_field} (a).
 \par Over-complete networks in deep learning is a new topic and was initially proposed for  medical image segmentation of small anatomy \cite{valanarasu2020kiu}. It has since been successfully extended to solve fine-context requiring tasks like fine edge segmentation of 3D volumes \cite{valanarasu2020kiu2}, deep subspace clustering \cite{valanarasu2021overcomplete}, MRI reconstruction \cite{guo2021overandunder}, adversarial defense against videos \cite{lo2020overcomplete} and image restoration problems like single image de-raining \cite{yasarla2020exploring}.

\section{Methodology}
\label{sec: method}
    \label{proposed_method}
    \begin{figure*}[tb!]
        \centering
        \includegraphics[width=0.6\linewidth]{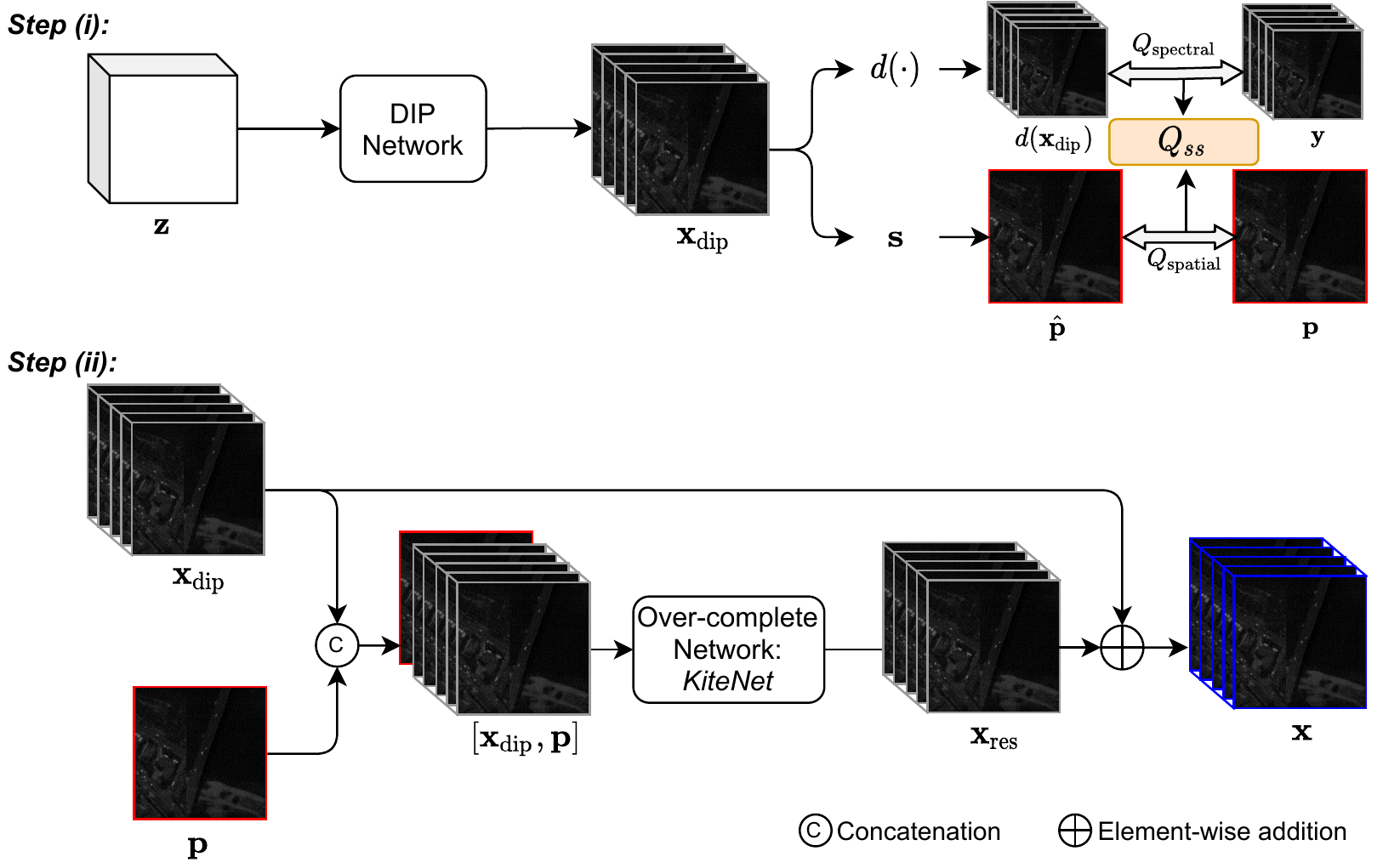}
        \caption{The overall flowchart of our proposed DIP-HyperKite for HS pansharpening. In the first step, we up-sample the LR-HSI $\mathbf{y}$ via DIP process to obtain the up-sampled HSI $\mathbf{x}_{\text{dip}}$. The DIP process takes a fixed noise tensor $\mathbf{z}$ as input for a given LR-HSI $\mathbf{y}$, and produces the up-sampled HSI $\mathbf{x}_{\text{dip}}$ by optimizing the proposed spatial+spectral energy function $Q_{\text{ss}}$ over the DIP network parameters $\theta$. In the second step, we take the up-sampled HSI $\mathbf{x}_{\text{dip}}$ and the PAN image $\mathbf{p}$ as inputs to predict the residual component $\mathbf{x}_{\text{res}}$ using our proposed over-complete network - HyperKite. Finally, the predicted residual image $\mathbf{x}_{\text{res}}$ is added to the up-sampled HSI $\mathbf{x}_{\text{dip}}$ to obtain the pansharpened HSI $\mathbf{x}$.} 
        \label{fig: proposed_method_flowchart}
    \end{figure*}
    \par The overall flowchart of the proposed DIP-HyperKite for HS pansharpening is shown in Figure \ref{fig: proposed_method_flowchart}. As can be seen from Figure \ref{fig: proposed_method_flowchart} the proposed method consists of two main steps. 
    In the first step, the LR-HSI $\mathbf{y} \in \mathbb{R}^{l \times w \times h}$ with $w \times h$ pixels and $l$ spectral bands is up-sampled to the spatial resolution of the PAN image $\mathbf{p} \in \mathbb{R}^{1 \times \beta w \times \beta h}$, where $\beta$ denotes the ratio between spatial resolution of $\mathbf{p}$ and $\mathbf{y}$. We denote the output from the DIP process as $\mathbf{x}_{\text{dip}} \in \mathbb{R}^{l \times \beta w \times \beta h}$. 
    In the second step, we train an over-complete deep network which takes up-sampled HSI $\mathbf{x}_{\text{dip}}$ and the corresponding PAN images $\mathbf{p}$ as inputs to predict the residual component $\mathbf{x}_{\text{res}}$ between the up-sampled HSI $\mathbf{x}_{\text{dip}}$ and the reference HSI $\mathbf{x}_{\text{ref}}$. 

\subsection{Up-sampling via DIP}
    \label{DIP}
    \par As shown in Figure \ref{fig: proposed_method_flowchart}, the low resolution HSI $\mathbf{y}$ is up-sampled to the spatial resolution of the PAN image $\mathbf{p}$ using the DIP. This recently introduced DIP method is different from the other existing up-sampling techniques such as bicubic interpolation, and LapSNR \cite{lap_snr}. The main advantage of DIP over these conventional methods is that it does not require a large dataset for training. In other words, for each LR image $\mathbf{y}$, the DIP network takes a fixed random tensor $\mathbf{z}$ as an input and optimize the network parameters $\theta$ by minimizing the loss function $Q$ which is defined in terms of the output up-sampled image $\mathbf{x}_{\text{dip}}$ and available LR-HSI $\mathbf{y}$ as given in (\ref{eq: standard_loss}). In contrast, the LapSRN network utilized in \cite{DDLPS} is highly relied upon the RGB image datasets and the knowledge adaptation techniques. Furthermore, the bicubic and LapSNR methods up-sample each band in the HSI separately; thus ignoring the high spatial correlation between the spectral bands, which results in the loss of spatial details. Although the DIP method is capable of producing high-quality upsampling images compared to the other existing methods, it only utilizes the information from the LR-HSI $\mathbf{y}$, thus only imposing constraint on the spectral domain. However, we observed that the quality of the sampled HSIs can be further improved by incorporating an additional spatial constraint in the loss function using the available PAN image $ \mathbf {p} $. In the next section we explain our novel spatial+spectral loss function. 
    
\subsubsection{Proposed spatial+spectral energy function for HS DIP}
    \label{proposed_dip}
    \begin{figure}[tb!]
        \centering
        \includegraphics[width=1\linewidth]{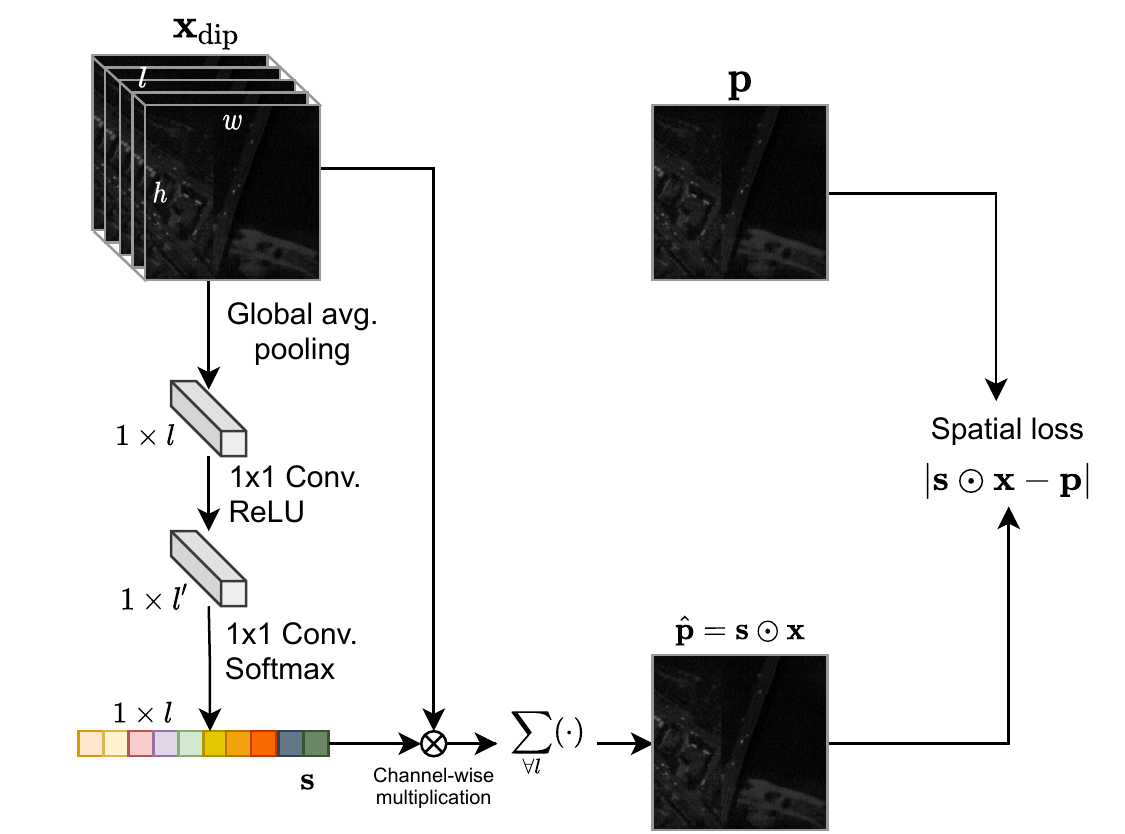}
        \caption{The proposed learnable spectral response function $\mathbf{s}$, and the computational procedure of evaluating the spatial loss term $Q_{\text{spatial}}$. We take the up-sampled HSI $\mathbf{x}_{\text{dip}}$ as the input, and feed it in to a Global Average Pooling (GAP) layer, which yielding a vector with a single entry for each spectral band. Then we pass it through a gating mechanism  by  forming  a  bottleneck  with  two  fully-connected (FC) layers ($1 \times 1$ convolutions) around the non-linearity to learn the spectral response of each band. Next, we apply a Softmax activation function to obtain \textit{normalized} spectral response $\mathbf{s}$, and then take the channel-wise multiplication followed by channel averaging to obtain the estimated PAN image $\hat{\mathbf{p}}$. Finally, we compute the the $L_1$ distance between the estimated PAN image $\hat{\mathbf{p}}$ and the reference PAN image $\mathbf{p}$ to obtain the spatial loss $Q_{\text{spatial}}$.}
        \label{fig:compute_spatial_loss}
    \end{figure}
    
    As we discussed in Section \ref{rw: dip}, the energy function given in (\ref{eq: standard_loss}) enforces a constraint only in spectral domain by defining the $L_1$ distance between up-sampled HSI $\mathbf{x}_{\text{dip}}$ and the LR HSI $\mathbf{y}$. Instead, we propose a loss function (denoted by $Q_{ss}$) for HS DIP, which enforces the constraints in both spatial and spectral domains as follows:
    \begin{equation}
        Q_{ss} = \underbrace{\norm{d(\mathbf{x}_{\text{dip}})-\mathbf{y}}_1}_{\text{spectral energy}} + \lambda \underbrace{\norm{ \left( \sum_{i \in \forall l} \mathbf{s}[i] \odot \mathbf{x}_{\text{dip}}[i] \right) - \mathbf{p}}_1}_{\text{spatial energy}},
        \label{eq: our_loss}
    \end{equation}
    where $\mathbf{s} \in \mathbb{R}^{1 \times l}$ denotes the spectral response function,  $\mathbf{s}[i]$ (scalar) is the spectral response of $i$-th band, $\mathbf{x}[i] \in \mathbb{R}^{h \times w}$ is  the $i$-th band image of the up-sampled HSI $\mathbf{x}_{\text{dip}}$, $\odot$ is the element-wise multiplication, and $\lambda$ is a regularization constant. The first term in (\ref{eq: our_loss}) enforces the spectral constraint on $\mathbf{x}$ as in (\ref{eq: standard_loss}), and the additional second term enforces the constraint in spatial domain on $\mathbf{x}_{\text{dip}}$ by utilizing the available PAN image $\mathbf{p}$. 
    \par In the simplest case, the spectral response function can be approximated as the average across all spectral bands (i.e. $\mathbf{s}[i] = 1/l; \forall i \in [1,l]$) \cite{SpectralRes1, SpectralRes2}. In this scenario, the spatial loss term in (\ref{eq: our_loss}) enforces that the average across all the spectral bands in up-sampled HSI $\mathbf{x}_{\text{dip}}$ to be close as possible to the PAN image $\mathbf{p}$, thus assuming a flat (i.e. uniform) spectral response. However, in general, this assumption is not valid as spectral response varies with wavelength coverage and different spectral bands describe the same semantic information across a wide spectral range with varying quality (i.e. PSNR) \cite{SpectralResEigen}. 
    
    \par A recent attempt \cite{SpectralResEigen} estimate the spectral response function $\mathbf{s}$ by utilizing the larger eigenvalue of the structure tensor (ST) matrix  (originally proposed in Harris corner detection algorithm  \cite{harris1988combined}). However, this method cannot be directly utilized in an end-to-end deep learning network due to the difficulties encountered while performing  back-propagation  In addition, it is highly computationally complex as it requires to compute derivatives of each band image along both $x$-and $y$-directions at each iteration of learning as part of constructing the structure tensor matrix. Instead, we propose a computationally lightweight and learnable spectral response function which can be easily integrated into the spatial loss term in (\ref{eq: our_loss}) and can be simultaneously learned with DIP.
    
    \par In this part we describe our novel way of estimating the spectral response function which is computationally lightweight, differentiable, and can be easily integrated into the existing DIP learning process. The overall computational procedure of estimating the spectral response function and thereby evaluating the spatial energy that we introduced for the DIP process in (\ref{eq: our_loss}) is graphically depicted in Figure \ref{fig:compute_spatial_loss}. First, we assume that the spectral response is proportional to the ratio of information in each spectral band. The next problem arises with this assumption is how do we quantify the information embedded in each spectral band. Motivated by recently proposed Squeeze-and-Excitation networks, we utilize global average pooling to quantify the global information present in each band. Formally, a statistic $\mathbf{q} \in \mathbb{R}^{1 \times l}$  which quantifies the informative features in each spectral band is generated by shrinking the up-sampled HSI $\mathbf{x}_{\text{dip}}$ through its spatial dimensions $h \times w$ such that the $i$-th element in $\mathbf{q}$ is calculated as:
    \begin{equation}
    	\mathbf{q}(i) = \frac{1}{h \times w} \sum_{\tilde{h} = 1}^{h} \sum_{\tilde{w} = 1}^{w} \mathbf{x}_{i}(\tilde{w}, \tilde{h}).
        \label{global_avg_pooling}
    \end{equation}
    Next, we use a simple gating mechanism to capture the dependencies among spectral bands using the band-wise descriptor $\mathbf{q}$ that we obtained in the previous step. We parameterize the spectral response function $\mathbf{s}$ by forming a bottleneck with two fully-connected (FC) layers around the non-linearity as follows:
    \begin{equation}
    	\mathbf{s} = \sigma (\mathbf{w}_2 \: \delta (\mathbf{w}_1 \mathbf{q})),
        \label{gating_mechanism}
    \end{equation}
    where $\sigma$ is the Sigmoid activation function, $\delta$ is the ReLU non-linearity, and $\mathbf{w}_1, \mathbf{w}_2$ are the learnable weight matrices. Here, we use Sigmoid activation to guarantee that the spectral responses of all the bands sump up to one.

\subsubsection{DIP network}
    \label{sec: DIP_network}
    \par
\begin{figure*}[tb]
        \centering
        \includegraphics[width=0.8\linewidth]{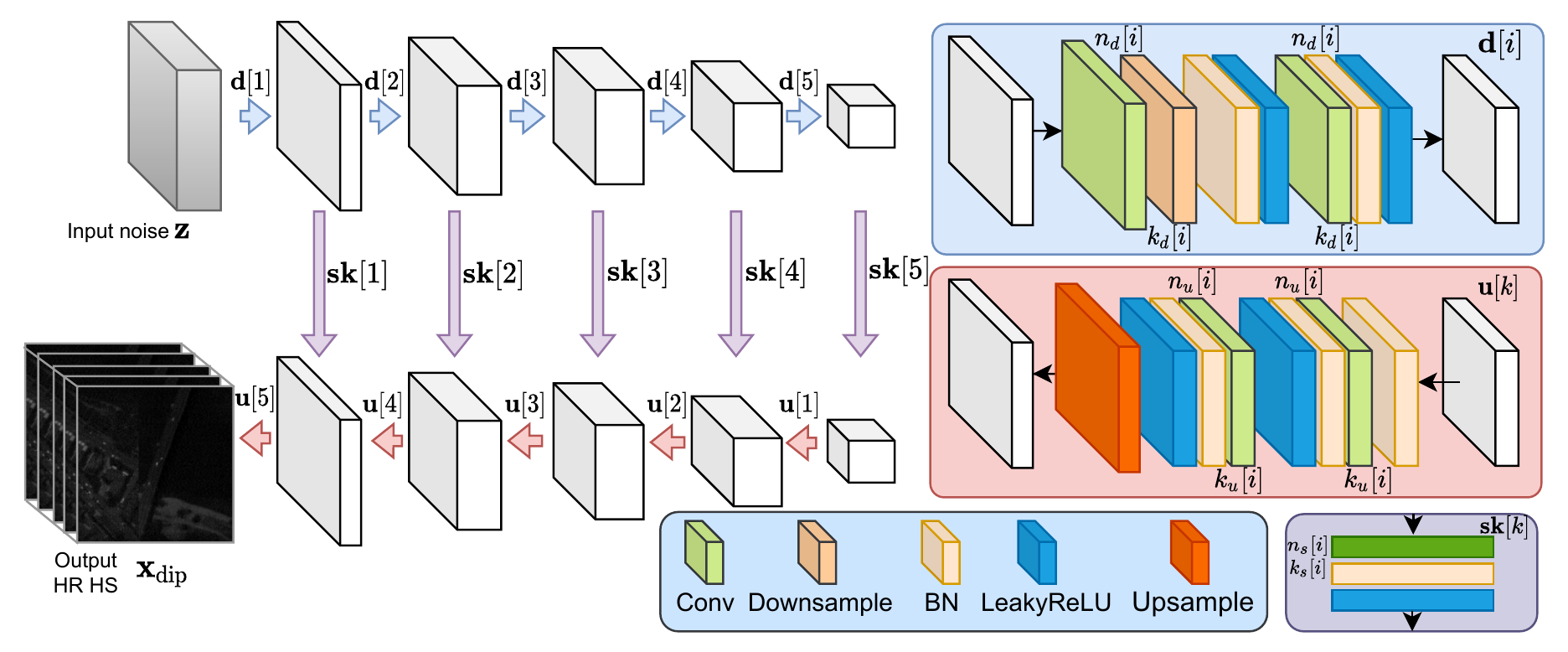}
        \caption{The DIP network utilized for the up-sampling process. The DIP network is a U-Net like network which consists of five down-sampling blocks $\mathbf{d}[i]$, five upsampling blocks $\mathbf{u}[i]$, and five skip-connection blocks $\mathbf{sk}[i]$ ($i=1,2,..,5$). The values of all the hyperparameters of DIP network is summarized in Table \ref{tab:dip_hyper}.}
        \label{fig:dhp_net}
    \end{figure*}

 Figure \ref{fig:dhp_net} illustrates a U-Net like deep network that we used for the DIP method. The DIP network includes five down-sampling blocks $\mathbf{d}[i]$, five upsampling blocks $\mathbf{u}[i]$, and five skip-connection blocks $\mathbf{sk}[i]$ ($i=1,2,..,5$). We use stride convolutions as the down-sampling operator, bi-linear up-sampling as the upsampling operator, and Lanczos2 as non-linearity.  We initialize the input noise vector with uniform noise between $0$ and $0.1$. The Table \ref{tab:dip_hyper} tabulates the values of all the hyperparameters of DIP network.
    
    \begin{table}[tb]
        \centering
        \caption{Hyperparameter values of the DIP network.}
        \begin{tabular}{ll}
            \hline
            Hyperparameter &  Value\\
            \hline
            $z$  & $\mathbb{R}^{32 \times \beta h \times \beta w} \sim U(0, 0.1)$\\
            $n_d = n_u$ & $[128, 128, 128, 128, 128]$\\
            $k_d = k_u$ & $[3, 3, 3, 3, 3]$\\
            $n_s$ & $[4, 4, 4, 4, 4]$\\
            $k_s$ & $[1, 1, 1, 1, 1]$\\
            Optimizer       & Adam\\
            Number of iterations        & 1300\\
            Learning rate   & 0.001\\
            Weight decay    & 0.0001\\
            Momentum        & 0.9\\
            Batch size      & 4\\
            LeakyReLU slope & 0.2\\
            \hline
        \end{tabular}
        \label{tab:dip_hyper}
    \end{table}
    
    \begin{figure}[tb]
        \centering
        \includegraphics[width=0.7\linewidth]{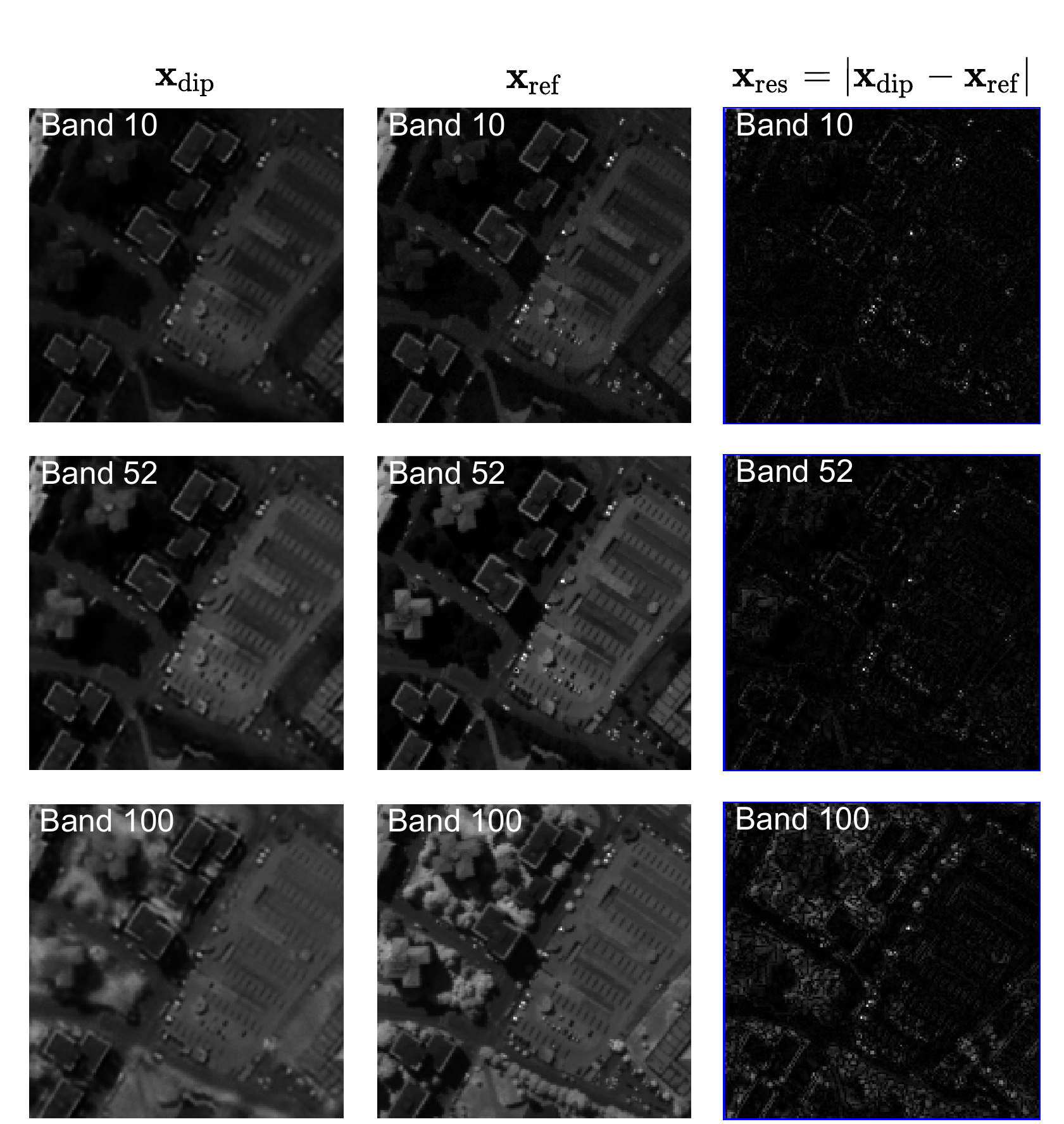}
        \caption{We observed that the residual component $\mathbf{x}_{\textbf{res}}$ (see third column) between up-sampled HSI $\mathbf{x}_{\text{dip}}$ (see first column) and the reference HSI $\mathbf{x}_{\textbf{ref}}$ (see second column) mainly consists of boundary information and very fine structures. To support this observation we show the residual component $\mathbf{x}_{\textbf{res}}$ for three different wavelength bands (i.e. band 10, band 52, and band 100) in the Pavia Center data set which will be introduced in Section \ref{sec:datasets}. 
        This observation motivated us to use an over-complete  network for the residual learning task, which is highly capable of learning low-level features such as fine edges and structures by transforming the input image into a higher dimension. We recommend that readers zoom in on this image to get a close-up view.}
        \label{fig: res_component}
    \end{figure}

\subsection{Residual learning via over-complete HyperKite}
    \label{sec:kiu-net}
    \begin{figure*}[tb]
        \centering
        \includegraphics[width=\linewidth]{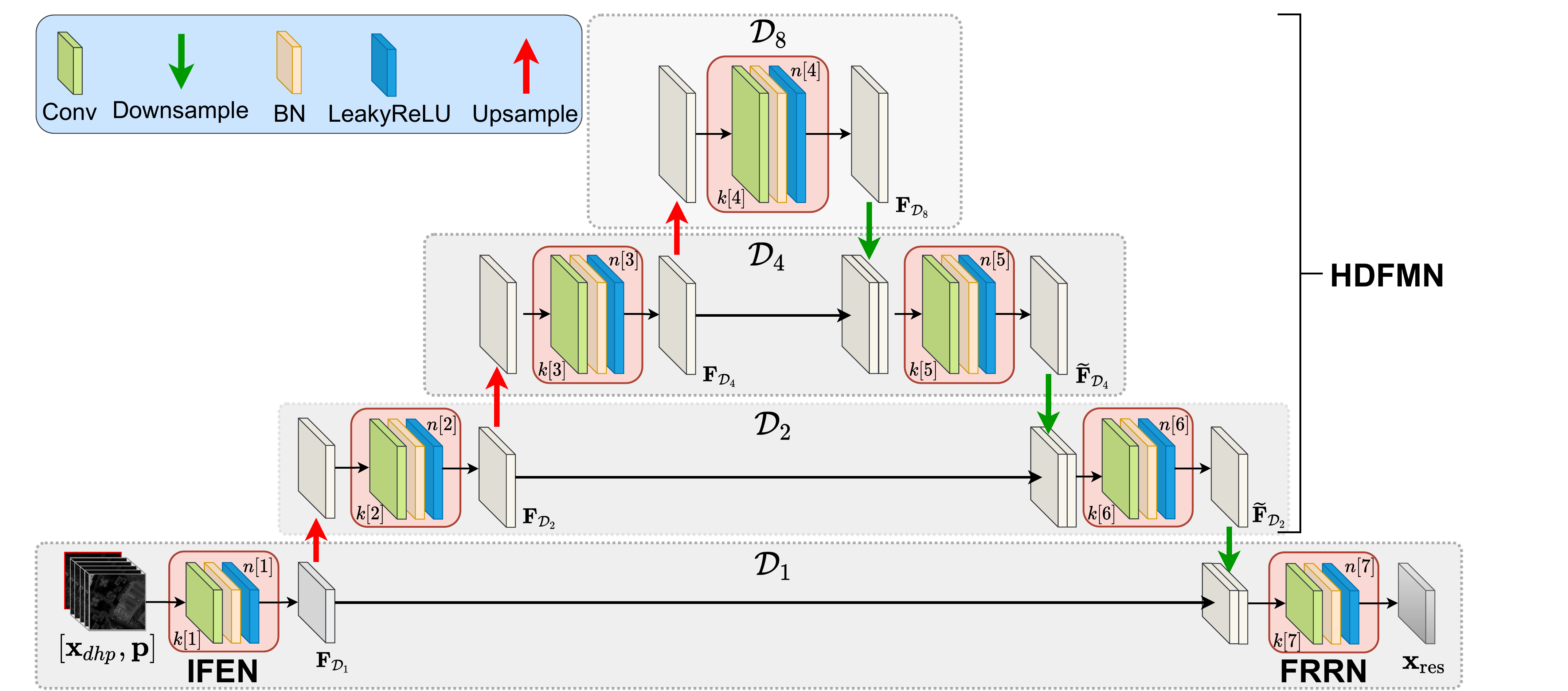}
        \caption{The proposed HyperKite architecture for the residual prediction task. We denote the  kernel size and the number of filters associated with each convolution block (shown in red color box) as $k[\cdot]$ and $n[\cdot]$, respectively. The values of all hyperparameters for  HyperKite is summarized in Table \ref{tab:HyperKite_hyper}.}
        \label{fig:HyperKite}
    \end{figure*}
    
    \begin{table}[tb]
        \centering
        \caption{Hyperparameter values of HyperKite}
        \begin{tabular}{ll}
            \hline
            Hyperparameter &  Value\\
            \hline
            $n$             & $[32, 64, 128, 128, 64, 32, l]$\\
            $k$             & $[3, 3, 3, 3, 3, 3, 3]$\\
            Optimizer       & Adam\\
            Num\_it         & 2500\\
            Learning rate   & 0.001\\
            Weight decay    & 0.0001\\
            Momentum        & 0.9\\
            Batch size      & 4\\
            LeakyReLU slope & 0.2\\
            \hline
        \end{tabular}
        \label{tab:HyperKite_hyper}
    \end{table}
    
    \par 
    Our motivation to design an over-complete network for the residual learning task emerged after observing the residual images between DIP up-sampled image $\mathbf{x}_{\text{dip}}$ and reference HSI $\mathbf{x}_{\text{ref}}$ as  visualized in Figure \ref{fig: res_component}. As we can see from Figure \ref{fig: res_component}, the residual images correspond to different wavelength band mainly consists of boundary information like edges and other high-frequency components. In order to accurately capture this fine information, we design an over-complete HyperKite for the residual learning as shown in Figure \ref{fig:HyperKite}.
    
\begin{figure}[tb]
        \centering
        \includegraphics[width=\linewidth]{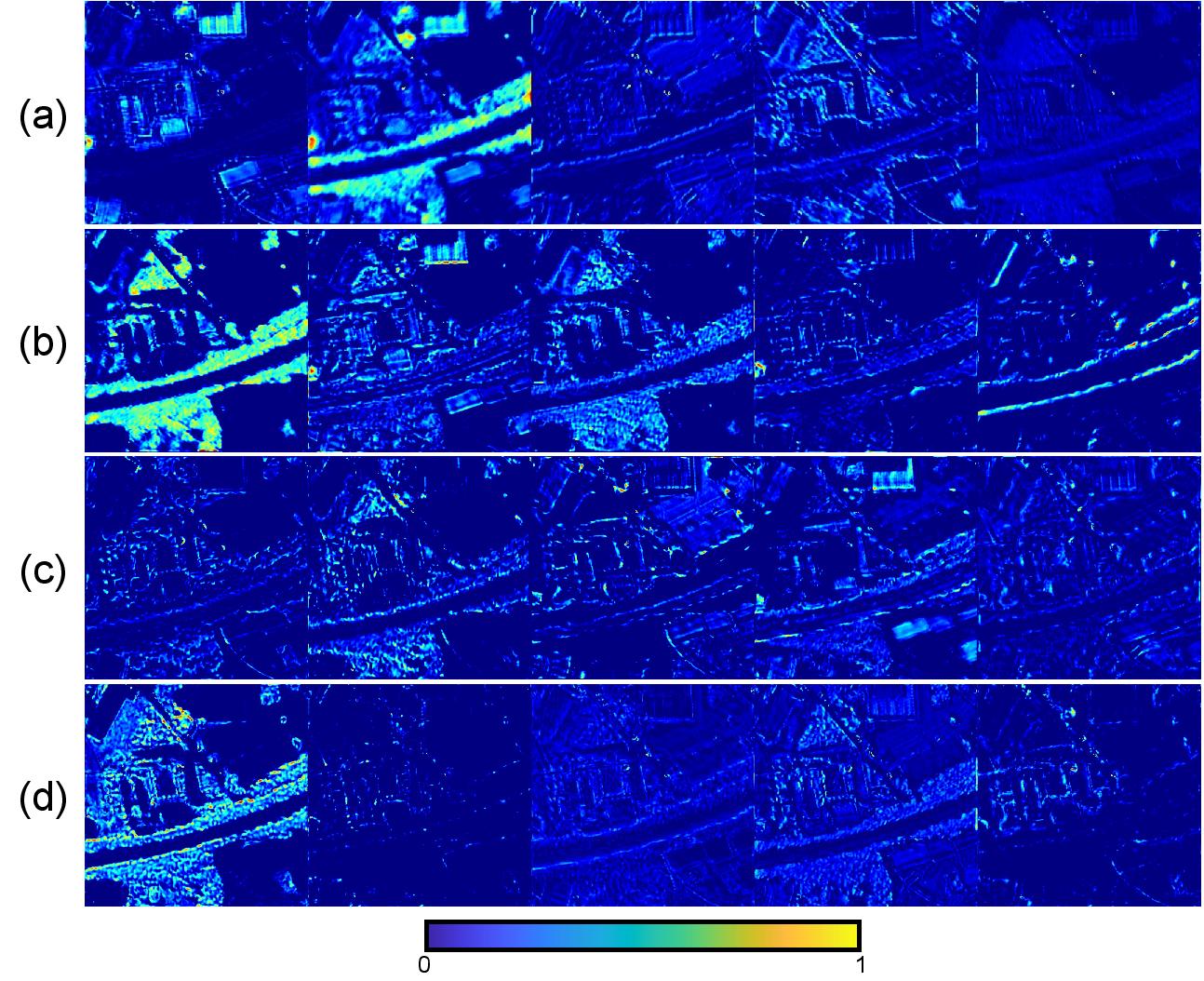}
        \caption{Visualization of filter responses of HyperKite. (a) Feature maps from the first layer of encoder. (b) Feature maps from the second layer of encoder. (c) Feature maps from the third layer of encoder. (d) Feature maps from the third layer of encoder. By restricting the receptive field, HyperKite is able to focus on edges and smaller regions. Zoom in recommended.}
        \label{fig:HyperKite_feature_maps}
    \end{figure}

    The proposed HyperKite consists of an Initial Feature Extraction Network (IFEN), a High-dimensional Feature Mapping Network (HDFMN), and a Final Residual Reconstruction Network (FRRN). The input to the HyperKite $\mathbf{x}_{\text{in}}$ is obtained by concatenating the up-sampled HSI $\mathbf{x}_{\text{dip}}$ and the PAN image $\mathbf{p}$ along the spectral dimension (denoted as $[\mathbf{x}_{\text{dip}}, \mathbf{p}]$). The HyperKite starts with the IFEN layer, where one $3 \times 3$ convolutional layer is applied followed by Batch Normalization (BN) and LeakyReLU non-linearity to extract initial feature representation as:
    \begin{equation}
        \mathbf{F}_{\mathcal{D}_1} = f_{\text{IFEN}}(\mathbf{x}_{\text{in}})
    \end{equation}
    where $f_{\text{IFEN}}(\cdot)$ denotes the $3\times3$ convolution followed by LeakyReLU and batch normalization, $\mathbf{F}_{\mathcal{D}_1}$ denotes the extracted features transformed from $\mathbf{x}_{\text{in}}$ in $\mathcal{D}_1 \in \mathbb{R}^{n[0] \times \beta w \times \beta h}$ dimensional pixel-space, and $n[0]$ is the number of filters in the convolutional layer. Figure \ref{fig:HyperKite_feature_maps} (a) shows six example feature maps of $\mathbf{F}_{\mathcal{D}_1}$ for the 20-th patch of the Pavia Center dataset that we will introduce in Section \ref{sec: experimental_settings}. As we can see from the figure, the initial feature extraction network $f_{\text{IFEN}}(\cdot)$ extract low-level feature of the input $\mathbf{x}_{\text{in}}$. In order to capture high-level features that that required for the residual learning, we successively transform the output of IFEN into three higher-dimensional pixel-spaces by utilizing the ``bilinear'' up-sampling denoted as $\mathcal{D}_2 \in \mathbb{R}^{2\beta w \times 2\beta h}$, $\mathcal{D}_4\in \mathbb{R}^{4\beta w \times 4\beta h}$, and $\mathcal{D}_8 \in \mathbb{R}^{8\beta w \times 8\beta h}$. Then we perform $3 \times 3$ convolution followed by BN and LeakyReLU to extract meaningful high-level features at each higher-dimensional space as, 
    \begin{align}
        \mathbf{F}_{\mathcal{D}_2} &= f_{\mathcal{D}_2} (\uparrow \mathbf{F}_{\mathcal{D}_1})\\
        \mathbf{F}_{\mathcal{D}_4} &= f_{\mathcal{D}_4} (\uparrow \mathbf{F}_{\mathcal{D}_2})\\
        \mathbf{F}_{\mathcal{D}_8} &= f_{\mathcal{D}_8} (\uparrow \mathbf{F}_{\mathcal{D}_4})
    \end{align}
    where $\uparrow$ denotes the ``bilinear'' interpolation by a factor of $2$, $f_{\mathcal{D}_{\text{d}}}(\cdot): \text{d} \in \{2,4,8\}$ denotes the $3 \times 3$ convolution layer followed by BN and LeakyReLU at the d-th higher-dimensional feature space. Next, we successively transform the extracted high-level features to the original dimensional space $\mathcal{D}_1$ by employing ``bilinear'' downsampling and skip connections. Formally, we can define the operations of HDFMN as,
    \begin{align}
        \widetilde{\mathbf{F}}_{\mathcal{D}_4} &= \widetilde{f}_{\mathcal{D}_4} (\downarrow \mathbf{F}_8 \oplus \mathbf{F}_4)\\
        \widetilde{\mathbf{F}}_{\mathcal{D}_2} &= \widetilde{f}_{\mathcal{D}_2} (\downarrow \widetilde{\mathbf{F}}_{\mathcal{D}_4} \oplus \mathbf{F}_2) \\
        \widetilde{\mathbf{F}}_{\mathcal{D}_1} &= \downarrow \widetilde{\mathbf{F}}_{\mathcal{D}_4},
    \end{align}
    where $\downarrow$ denotes the ``bilinear'' downsampling by a factor of $2$, $\oplus$ denotes the feature concatenation operator, $f_{\mathcal{D}_{\text{d}}}(\cdot): \text{d} \in \{0, 2, 4\}$ denotes the $3 \times 3$ convolution followed by BN and LeakyReLU at the d-th dimensional feature space, and $\widetilde{\mathbf{F}}_{\mathcal{D}_{\text{d}}}$ is the most relevant high-level features obtained at $\mathcal{D}_{\text{d}} \in \{0, 2, 4\}$ space. After flowing through all the downsampling layers (decoder blocks), a $3 \times 3$ convolutional layer is employed to recover the spectral dimension, and reconstruct the residual image $\mathbf{x}_{\text{red}}$ as:
    \begin{align}
        \mathbf{x}_{\text{res}} &= f_{\text{FRNN}} (\widetilde{\mathbf{F}}_{\mathcal{D}_1} \oplus \mathbf{F}_{\mathcal{D}_1})
    \end{align}
    where $f_{\text{FRNN}}$  denotes the $3 \times 3$ convolutional layer followed by BN and LeakyReLU employed at FRNN. 
    
    \par After carrying out DIP up-sampling and residual prediction of our DIP-HyperKite, the DIP up-sampled HSIs $\mathbf{x}_{\text{dip}}$ and $\mathbf{x}_{\text{res}}$ are created. Finally, we can obtain the fused HSI $\mathbf{x}$ by using  $\mathbf{x}_{\text{dip}}$ and $\mathbf{x}_{\text{res}}$ as:
    \begin{equation}
        \mathbf{x} = \mathbf{x}_{\text{res}} + \mathbf{x}_{\text{dip}}. 
    \end{equation}
    
    \par To this end, we utilize $L_1$ loss to optimize HyperKite, which has been demonstrated as a superior choice for remote sensing image SR \cite{Hyper-PNN, hyperspectral_pan_review} and also experimentally verified to be effective for improving the fusion accuracy. For the training set $\{\mathbf{x}_{\text{in}}^k, \mathbf{x}_{\text{ref}}^k\}^L$  where $\mathbf{x}_{\text{in}}^k$ is the $k$-th input, $\mathbf{x}_{\text{ref}}^k$ is the corresponding reference HSI, and $L$ is the total number of training HSIs in the training set. The $L_1$ loss function utilized for HyperKite training can be defined as follows:
    \begin{equation}
        L(\theta) = \frac{1}{L} \sum_{k=1}^{L} \sum_{k=1}^{L} \norm{f_{\text{HyperKite}}(\mathbf{x}_{\text{in}}^k) - \mathbf{x}_{\text{res}}^k}_1.
    \end{equation}
    
   Moreover, all the parameter details of our proposed HyperKite are  summarized in Table \ref{tab:HyperKite_hyper}. We train our network in Pytorch framework using an NVIDIA Quadro 8000 GPU. We use Adam optimizer with a learning rate of 0.001, weight decay of 0.0001 and momentum 0.9 to train HyperKite. We use a batch size of 4 and train the network for 2500 epcochs.

\section{Experimental Settings}
\label{sec: experimental_settings}
    \subsection{Datasets}
        \label{sec:datasets}
        To evaluate the performance of our proposed DIP-HyperKite for HS pansharpening, we conduct a series of experiments on three HS data sets, which are described in detail below.
        
        \subsubsection{Pavia Center dataset} The Pavia Center scene was captured by the ROSIS camera \cite{ROSIS_camera}. The original HSI consists of $115$ spectral bands spanning from $430$ to $960$ nm. The spatial size of the original image is $1096 \times 1096$ pixels, where a single pixel is equivalent to geometric resolution of $1.3\times1.3$ m$^2$. The thirteen noisy spectral bands in the original HSI were discarded, thus resulting in a HSI with $102$ spectral bands spanning from $430$ to $860$ nm. In addition, a rectangular area of size $1096 \times 381$ pixels with no information at the center of the original HSI was also discarded, and the resulting ``two-part'' image with size of $1096 \times 715 \times 102$ was used for the experiments. Following the same experimental procedure outlined in \cite{DHP-DARN}, we also used only the top-left corner of the HSI with size of $960 \times 640 \times 102$, and partitioned it into $24$ cubic patches of size $160 \times 160 \times 102$ with no overlap, which constituted the reference images ($\mathbf{x}_{\text{ref}}$) of  Pavia Center data set. In order to generate PAN images ($\mathbf{p}$) and LR-HSIs ($\mathbf{y}$) corresponding to each HR-HSI, we utilize Wald's protocol \cite{walds_protocol}. Following the Wald's protocol, we generate PAN images ($\mathbf{p}$) of size $160 \times 160$ by averaging first $61$ spectral bands of HR reference HSI. In order to generate LR-HSIs of size $40 \times 40 \times 102$, we spatially blurred the HR reference HSI with an $8 \times 8$ Gaussian filter, and then downsampled the result. The scaling factor $(\beta)$ was set to $4$ for the Pavia Center dataset. We randomly select $17$ cubic patches for the training, and the rest of the seven patches forms the testing set of the Pavia Center dataset.
        
        \subsubsection{Botswana dataset} The Botswana scene was acquired by the Hyperion sensor on the NASA's Earth Observing 1 (EO-1) satellite. The original Botswana HSI consists of $242$ spectral bands spanning from $400$ to $2500$ nm with spectral resolution of $10$ nm. The spatial size of the original Botswana image is $1496 \times 256$ pixels. We remove the uncalibrated and noisy spectral bands in the original image, thus resulting in a HSI with $145$ spectral bands. Following the same experimental procedure outlined in \cite{DHP-DARN}, we also use only the top-left corner of the HSI with size of $1200 \times 240 \times 145$, and partitioned it into 20 cubic patches of size $120 \times 120 \time 145$ with no overlap, which constitute the reference images $\mathbf{x}_{\text{ref}}$ of the Botswana dataset. In order to generate PAN images $\mathbf{p}$ and the LR-HSIs $\mathbf{y}$ corresponding to each HR reference image, we follow the Wald's protocol. We generate PAN images $\mathbf{p}$ of size $120 \times 120$ by averaging first $31$ spectral bands of HR-HSI. To generate LR-HSIs $\mathbf{y}$, we spatially blur the HR-HSI with an $8 \times 8$ window, and perform  down-sampling. For the Botswana dataset, we set the down-sampling factor $\beta$ to 3. We randomly select $14$ cubic patches for  training, and the rest of the patches are utilized for  testing.
        
        \subsubsection{Chikusei dataset \cite{Chikusei_dataset}} The Chikusei scene was captured by the Headwall Hyperspec-VNIR-C imaging sensor over the agricultural and urban areas in Chikusei, Japan. The original Chikusei HSI consists of 128 spectral bands spanning from $363$ to $1018$ nm. The spatial size of the Chikusei HSI is $2517 \times 2335$ pixels, where a single pixel is equivalent to geometric resolution of $2.5 \times 2.5 $ m$^2$. We used top-left corner of the HSI with size of $2304 \times 2304 \times 128$, and partitioned it into $81$ cubic patches of size $256 \times 256 \times 128$ with no overlap, which constituted the reference images $\mathbf{x}_{\text{ref}}$ of Chikusei dataset. Following the Wald's protocol, we generate PAN images of size $256 \times 256$ by averaging first $65$ spectral bands of high resolution HSI. To generate LR-HSIs $\mathbf{y}$, we spatially blur the HR-HSI with an $8 \times 8$ window, and perform  down-sampling. For the Chikusei dataset, we set the down-sampling factor $\beta$ to $4$. We randomly select 61 cubic patches for training, and the rest of the patches are utilized for  testing.
        
        \paragraph*{Note} The standard deviation $(\sigma)$ of the Gaussian filter that we use to generate LR-HSIs is calculated as $\sigma = 0.4247 \beta $ \cite{Panshaperpening_wei_thesis}. 
        
    \subsection{Performance measures}
    In order to evaluate the quality of the proposed pansharpening method, we use different image quality measures. Following \cite{DHP-DARN}, we use Cross-Correlation (CC), Spectral Angle Mapping (SAM), Root Mean Square Error (RMSE), Errur Relative Globale Adimensionnelle Desynthese (ERGAS), and Peak Signal to Noise Ratio (PSNR). These measures have been widely used in the HSI processing community and are appropriate for evaluating  fusion in spectral and spatial resolutions.
    
    \subsubsection{Cross-Correlation (CC)} The CC metric characterizes the geometric distortion, and is defined as:
    \begin{equation}
        \text{CC}(\mathbf{x}, \mathbf{x}_{\text{ref}}) = \frac{1}{l} \sum_{i=1}^{l} \text{CCS}(\mathbf{x}^{i}, \mathbf{x}_{\text{ref}}^{i})
    \end{equation}
    where, CCS denotes the cross-correlation for a single-band image as follows:
    \begin{equation}
        \text{CCS}(\mathbf{A}, \mathbf{B}) = \frac{\sum_{j=1}^{n} (\mathbf{A}_j - \mu_A)(\mathbf{B}_j - \mu_B)}{\sqrt{\sum_{j=1}^{n} (\mathbf{A}_j - \mu_A)^2\sum_{j=1}^{n}(\mathbf{B}_j-\mu_B)^2}}
    \end{equation}
    where, $\mu_A = \frac{1}{n} \sum_{j=1}^{n} \mathbf{A}_j$ is the sample mean of $\mathbf{A}$. The ideal value of CC is $1.0$, which indicates that the two HSIs are highly correlated.
    
    \subsubsection{SAM}  SAM is a spectral measure which is defined as:
    \begin{equation}
        \text{SAM}(\mathbf{x}, \mathbf{x}_{\text{ref}}) = \frac{1}{n} \sum_{j=1}^{n} \text{SAM}(\mathbf{x}_j, \mathbf{{x}_{\text{ref}}}_j),
    \end{equation}
    where given the vectors $\mathbf{a}$, $\mathbf{b} \in \mathbb{R}^{l}$,
    \begin{equation}
        \text{SAM}(\mathbf{a}, \mathbf{b}) = \arccos{\left( \frac{<\mathbf{a}, \mathbf{b}>}{\lVert \mathbf{a}\rVert \lVert\mathbf{b}\rVert} \right)},
    \end{equation}
    where $<\mathbf{a}, \mathbf{b}>$ denotes the inner product between $\mathbf{a}$ and $\mathbf{b}$, and $\lVert \cdot \rVert$ is the $L_2$ norm. The SAM is a measure of the spectral shape preservation. The SAM values reported in our experiments are in degrees and thus belongs to $(-90, 90]$. The optimal value of SAM is $0.0$. The values of SAM reported in our experiments have obtained by averaging the values for all image pixels.
    
    \subsubsection{RSNR/ RMSE} The reconstruction SNR (RSNR) or root mean square error (RMSE) is related to the difference between the reference and fuse images, which is defined as follows:
    \begin{equation}
        \text{RMSE}(\mathbf{x}, \mathbf{x}_{\text{ref}}) = \frac{1}{n \times l} \rVert \mathbf{x} - \mathbf{x}_{\text{ref}} \lVert^2_F,
    \end{equation}
    \begin{equation}
        \text{RSNR}(\mathbf{x}, \mathbf{x}_{\text{ref}}) = 10 \log_{10} \left( \frac{\lVert \mathbf{x}_{\text{ref}} \rVert^2_F}{ \lVert \mathbf{x} - \mathbf{x}_{\text{ref}}  \rVert^2_F  } \right).
    \end{equation}
    
    \subsubsection{ERGAS} Relative dimensionless global error in synthesis (ERGAS) calculates the amount of spectral distortion in the image. The ERGAS measure is defined as:
    \begin{equation}
        \text{ERGAS} = 100 \frac{1}{d^2} \sqrt{\frac{1}{l} \sum_{i=1}^{l} \left( \frac{\text{RMSE}(i)}{\mu_i} \right)  },
    \end{equation}
    where $d$ is the ratio between the linear resolution of the PAN image and the HSIs. defined as:
    \begin{equation}
        d = \frac{\text{PAN linear spatial resolution}}{\text{HS linear spatial resolution}},
    \end{equation}
    $\text{RMSE}_i = \frac{\lVert \mathbf{x}^i - \mathbf{x}_{\text{ref}}^i \rVert_F}{\sqrt{n}}$, and $ \mu_i$ i the sample mean of the $i$-th band of $\mathbf{x}_{\text{ref}}$. The ideal value of ERGAS is 0.
    
    \subsubsection{Peak Signal to Noise Ratio (PSNR)} PSNR also assess the fusion quality of each bans, and the average PSNR is calculated as:
    \begin{equation}
        \text{PSNR} = \frac{1}{l} \sum_{i=1}^{l} \left[ 10 \log_{10} \left(  \frac{\max \left( \mathbf{x}_{\text{ref}}^i \right)}{\text{RMSE}_i}  \right)^2 \right]
    \end{equation}
    where $\max \left( \mathbf{x}_{\text{ref}}^i \right)$ is the maximum pixel value in the $i$th band of $\mathbf{x}_{ref}$. A larger value of PSNR indicates a higher reconstruction quality in spatial information of the fusion result.

\section{Results and Discussion}
\label{sec: results}
    This section presents the results of our proposed DIP-HyperKite for HS pansharpening, and compares it with the state-of-the-art methods on the Pavia Center, Botswana, and Chikusei datasets. For better clarity, we divide this section into two parts. In the first part (section \ref{results_upsample}), we highlight the contribution from our proposed spatial+spectral energy function for the DIP up-sampling process and compare it with available state-of-the-art up-sampling techniques such as  nearest-neighbor, bicubic, LapSRN, and DIP with only spectral loss. In the second part (section \ref{sec:final_fusion_results}), we present the final fusion results that we obtain from our proposed HyperKite network and compare it with classical and deep-learning-based pansharpening approaches. 
    \begin{figure}[tb]
            \centering
            \includegraphics[width=\linewidth]{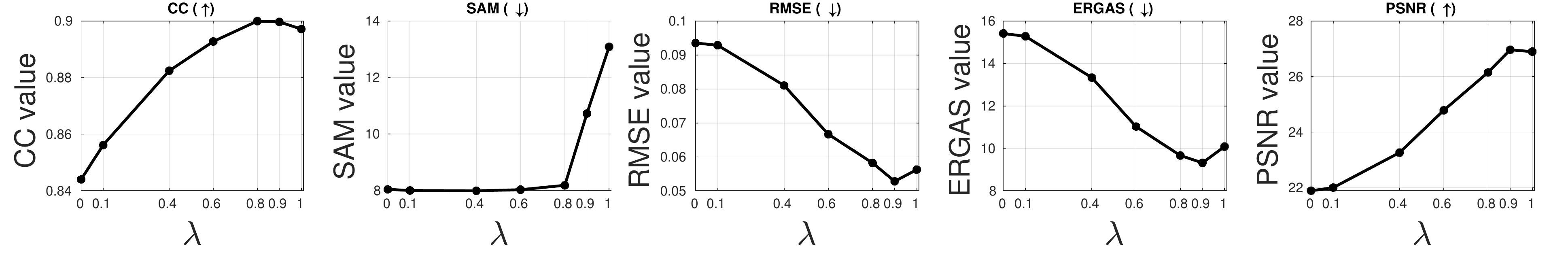}
            \caption{The variation of CC, SAM, RMSE, ERGAS, and PSNR with the regularization constant $\lambda$ in our spectral+spectral energy function $Q_{ss}$ for Pavia Center dataset. We select $\lambda= 0.8$ as the optimal value of regularization constant for the Pavia Center dataset by considering all the performance metrics.}
            \label{fig:pavia_lambda_tune_ql}
    \end{figure}
    \begin{figure}[tb]
        \centering
        \includegraphics[width=\linewidth]{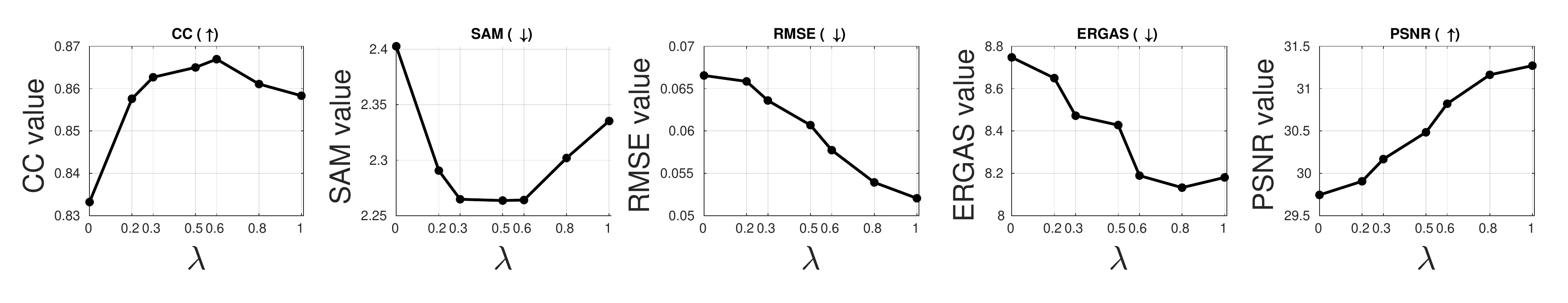}
        \caption{The variation of CC, SAM, RMSE, ERGAS, and PSNR with the regularization constant $\lambda$ in our spectral+spectral energy function $Q_{ss}$ for Botswana dataset. We select $\lambda= 0.8$ as the optimal value of regularization constant for the Botswana dataset by considering all the performance metrics.}
        \label{fig:botswana_lambda_tune_ql}
    \end{figure}
    \begin{figure}[tb]
        \centering
        \includegraphics[width=\linewidth]{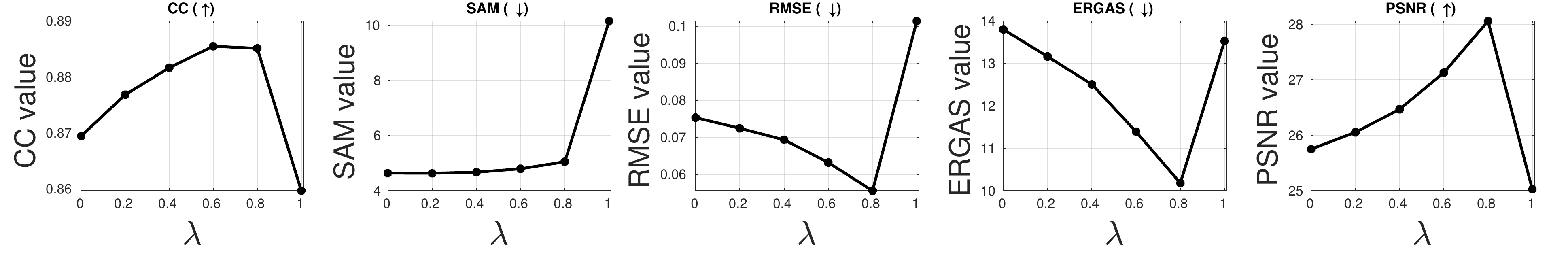}
        \caption{The variation of CC, SAM, RMSE, ERGAS, and PSNR with the regularization constant $\lambda$ in our spectral+spectral energy function $Q_{ss}$ for Chikusei dataset. We select $\lambda= 0.8$ as the optimal value of regularization constant for the Chikusei dataset by considering all the performance metrics.}
        \label{fig:chikusei_lambda_tune_ql}
    \end{figure}
    
    \subsection{Effect of the proposed spatial+spectral energy function for the DIP up-sampling process} 
    \label{results_upsample}
    As we discussed in Section \ref{DIP}, the recently proposed pansharpening methods such as DHP-DARN \cite{DHP-DARN} and DHP \cite{DeepHyperspectralPrior} utilized the DIP process to up-sample the LR-HSI instead of using the nearest-neighbor, bicubic, or LapSRN techniques due to its excellent performance. However, we have observed that the quality of up-sampled HSI can be further improved by carefully redesigning the loss function used in the DIP optimization. Instead of only utilizing spectral constraint in the DIP loss function, we derived a novel loss function with spectral and spatial constraints. This section demonstrates the performance improvement brought by our proposed spatial+spectral loss function to the DIP up-sampling process. We compare DIP with the  proposed spatial+spectral loss against the DIP with spectral loss only. Furthermore, to make the analysis more comprehensive,  we also added a conventional up-sampling techniques used in the HS pansharpening domain, such as nearest-neighbor, and bicubic. Further, motivated by the experimental discussion in \cite{DHP-DARN}, we also added the results from LapSRN \cite{LapSRN}, which is trained on a large amount of RGB images.

    \begin{table}[tb]
        \centering
        \caption{Average quantitative results for different up-sampling techniques on the Pavia Center dataset.}
        \begin{tabular}{lp{0.5cm}p{0.5cm}p{0.7cm}p{0.6cm}p{0.7cm}p{0.6cm}}
        \hline
        \multirow{3}{*}{Method} & CC   & SAM & RMSE  & RSNR   & ERGAS       & PSNR \\
        & & & $\times 10^{-1}$ & & & \\
        & ($\uparrow$) & ($\downarrow$) & ($\downarrow$) & ($\uparrow$) & ($\downarrow$) & ($\uparrow$)\\
        \hline
        Nearest-neighbor    & 0.809 & 7.70  & 1.22  & 9.63  & 19.97     & 19.65\\
        Bicubic             & 0.840 & 7.45  & 1.13  & 11.26 & 18.48     & 20.36\\
        LapSRN \cite{LapSRN}& 0.843 & \textbf{7.37} & 1.12 & 11.56 & 18.16 & 20.49\\
        DIP+spectral\cite{DHP-DARN} & 0.844  & 8.04   & 0.94   & 14.91 & 15.42 & 21.89\\
        DIP+$Q_{\text{ss}}$ (ours)& \textbf{0.900}   & 8.18            & \textbf{0.58}     & \textbf{24.36}& \textbf{9.66}       & \textbf{26.15}\\
                            & \textbf{\tiny(+6.6\%)} & \textbf{-} & \textbf{\tiny (-37.8\%)} &    \textbf{\tiny (+63.4\%)} & \textbf{\tiny (-37.3\%)} & \textbf{\tiny (+19.5\%)} \\
        \hline
        \end{tabular}
        \label{tab:pavia_lambda_tune_qt}
    \end{table}
    \begin{table}[tb]
        \centering
        \caption{Average quantitative results for different up-sampling techniques on the Botswana dataset.}
        \begin{tabular}{lp{0.5cm}p{0.6cm}p{0.7cm}p{0.6cm}p{0.7cm}p{0.6cm}}
        \hline
        \multirow{3}{*}{Method} & CC   & SAM & RMSE  & RSNR   & ERGAS       & PSNR \\
        & & & $\times 10^{-2}$ & & & \\
        & ($\uparrow$) & ($\downarrow$) & ($\downarrow$) & ($\uparrow$) & ($\downarrow$) & ($\uparrow$)\\
        \hline
        Nearest-neighbor    & 0.854 & 2.52  & 7.87  & 29.03 & 9.08 & 28.87\\
        Bicubic             & 0.852 & 2.42  & 7.67  & 29.60 & 8.77 & 29.17\\
        LapSRN \cite{LapSRN}& 0.858 & 2.47  & 6.27  & 34.01 & 8.27 & 29.01\\
        DIP+spectral  \cite{DHP-DARN}  & 0.833 & 2.40  & 6.66   & 32.91 & 8.75  & 29.75\\
        DIP+$Q_{\text{ss}}$ (ours)& \textbf{0.861}   & \textbf{2.30}   & \textbf{5.39}     & \textbf{37.80}    & \textbf{8.13}       & \textbf{31.28}\\
                            & \textbf{\tiny(+0.4\%)} &  \textbf{\tiny(-4.2\%)}          & \textbf{\tiny(-14.0\%)} &    \textbf{\tiny(+11.2\%)} & \textbf{\tiny(-1.7\%)} & \textbf{\tiny(+5.2\%)} \\
        \hline
        \end{tabular}
        \label{tab:botswana_lambda_tune_qt}
    \end{table}
    \begin{table}[tb]
        \centering
        \caption{Average quantitative results for different up-sampling techniques on the Chikusei dataset.}
        \begin{tabular}{lp{0.5cm}p{0.6cm}p{0.7cm}p{0.6cm}p{0.7cm}p{0.6cm}}
        \hline
        \multirow{3}{*}{Method} & CC   & SAM & RMSE  & RSNR   & ERGAS       & PSNR \\
        & & & $\times 10^{-2}$ & & & \\
        & ($\uparrow$) & ($\downarrow$) & ($\downarrow$) & ($\uparrow$) & ($\downarrow$) & ($\uparrow$)\\
        \hline
        Nearest-neighbor    & 0.861 & 4.05  & 9.99 & 18.26 & 17.03  & 23.73\\
        Bicubic             & 0.884 & 3.86 & 0.093 & 20.07 & 15.75  & 24.52\\
        LapSRN \cite{LapSRN}& 0.885 & \textbf{3.75}& 8.53  & 21.37  & 14.33 & 25.06\\
        DIP+spectral\cite{DHP-DARN}   & 0.869  & 4.64   & 7.54  & 24.16 & 13.80 & 25.75\\
        DIP+$Q_{\text{ss}}$(ours)  & \textbf{0.885}   & 5.05 & \textbf{5.56} & \textbf{29.69}    & \textbf{10.18}  & \textbf{28.06}\\
                            & \textbf{\tiny(+0.1\%)} &         \textbf{-}          & \textbf{\tiny(-26.3\%)} &    \textbf{\tiny(+22.9\%)} & \textbf{\tiny(-26.2\%)} & \textbf{\tiny(+8.9\%)} \\
        \hline
        \end{tabular}
        \label{tab:chikusei_lambda_tune_qt}
    \end{table}
    \subsubsection{Tuning the hyperparameter $\lambda$ in our spatial+spectral energy function} 
    \par We start our discussion with the effect of the regularization constant $\lambda$ in our proposed spatial+spectral loss function as defined in (\ref{eq: our_loss}). The variation of CC, SAM, RMSE, ERGAS and PSNR values when varying the regularization parameter $\lambda$ from $0.0$ to $1.0$ for the Pavia Center, Botswana and Chikusei datasets are shown in Figure \ref{fig:pavia_lambda_tune_ql}, Figure \ref{fig:botswana_lambda_tune_ql}, and Figure \ref{fig:chikusei_lambda_tune_ql}, respectively. As can be seen from these figures, as the value of the regularization constant $\lambda$ increases, the performance metrics also begin to improve, then hit a saturation point, and then degrade, for all three data sets. Therefore, we carefully select the regularization constant $\lambda$ for each dataset by considering all the performance metrics. For example, consider the variation of the performance metrics with the regularization constant $\lambda$ for the Pavia Center dataset which is shown in Figure \ref{fig:pavia_lambda_tune_ql}. As we can see, when the value of the regularization constant increases from $0.0$ to $0.8$, we can see that CC, RMSE, ERGAS, and PSNR start to improve, and when $\lambda$ increases beyond $0.8$ the performance metrics start to degrade. Therefore, we set $\lambda = 0.8$ as the optimal value of the regularization constant of our proposed spatial+spectral energy term for the Pavia Center dataset. The variation in performance metrics with the regularization parameter $\lambda$ for the Botswana and Chikusei datasets are also shown in Figure \ref{fig:botswana_lambda_tune_ql} and Figure \ref{fig:chikusei_lambda_tune_ql}, respectively. Following the same analysis we described for the Pavia Center dataset, we select $\lambda = 0.8$ as the optimal value of the regularization constant for the Botswana and Chikusei datasets. Note that the performance improvement bringing from our proposed spatial+spectral loss function for the DIP upsampling process. Under the optimal regularization constant ($\lambda = 0.8$), our spatial+spectral energy function improves the quality of up-sampled HSIs over the spectral loss (equivalent to $\lambda = 0$ point in Figure \ref{fig:pavia_lambda_tune_ql}, \ref{fig:botswana_lambda_tune_ql}, and \ref{fig:chikusei_lambda_tune_ql}) in-terms of CC, RMSE, ERGAS, and PSNR metrics by 6.64\%,  37.8\%, 63.4\%, 37.3\%, and 19.5\%, respectively for the Pavia Center dataset. For the Botswana dataset, our proposed loss function improves CC, SAM, RMSE, ERGAS, and PSNR metrics over the DIP with spectral loss by 3.3\%,  4.2\%, 19.1\%, 14.7\%, 7.0\%, and $5.2\%$, respectively. Similarly for the Chikusei dataset, our method improves CC, RMSE, ERGAS, and PSNR metrics compared to  DIP with spectral loss by 1.8\%,  26.3\%, 22.9\%, 26.2\%, and 8.9\%, respectively.
    
    \begin{figure}[tb]
        \centering
        \includegraphics[width=\linewidth]{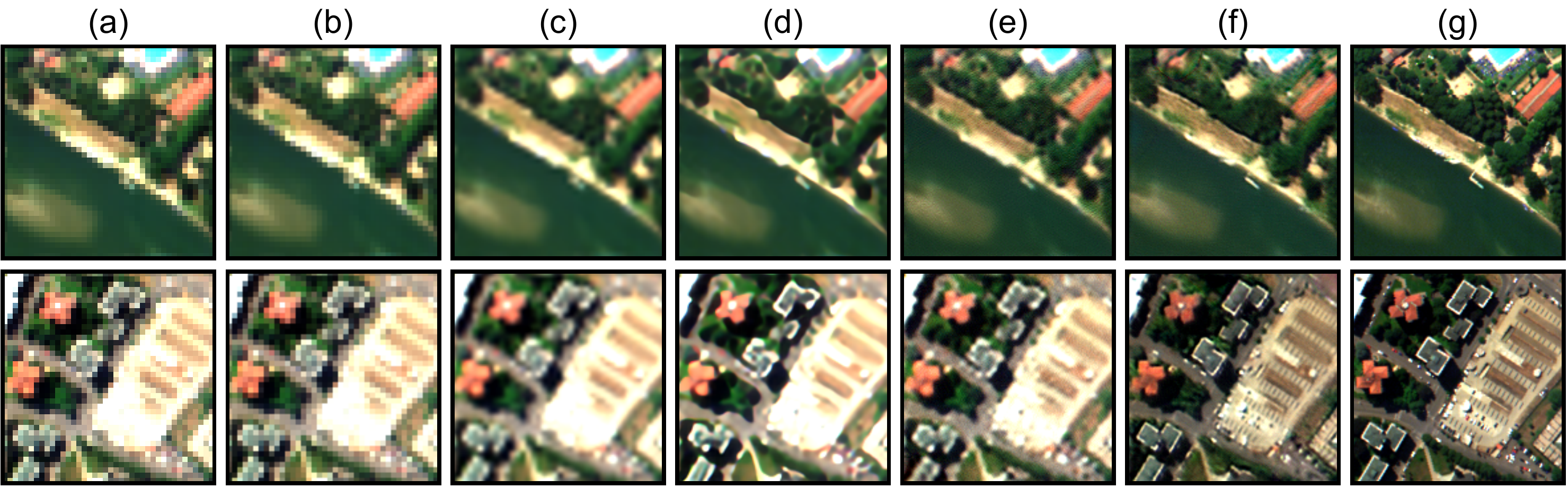}
        \caption{Up-sampled images of 1-st patch (in 1-st row) and 11-th patch (in 2-nd row) of Pavia Center dataset. (a) LR-HSI. (b) Nearest-neighbor. (c) Bicubic. (d) LapSRN \cite{LapSRN}. (e) DIP with only spectral energy \cite{DHP-DARN}. (f) DIP with our spatial+spectral energy $(Q_{ss}; \lambda=0.8)$. (g) Reference. The RGB image is generated by utilizing the 10-th, 30-th, and 60-th bands of the HSI for blue, green and red bands, respectively.}
        \label{fig:pavia_upsampling_comp}
    \end{figure}
    
    \begin{figure}[tb]
        \centering
        \includegraphics[width=\linewidth]{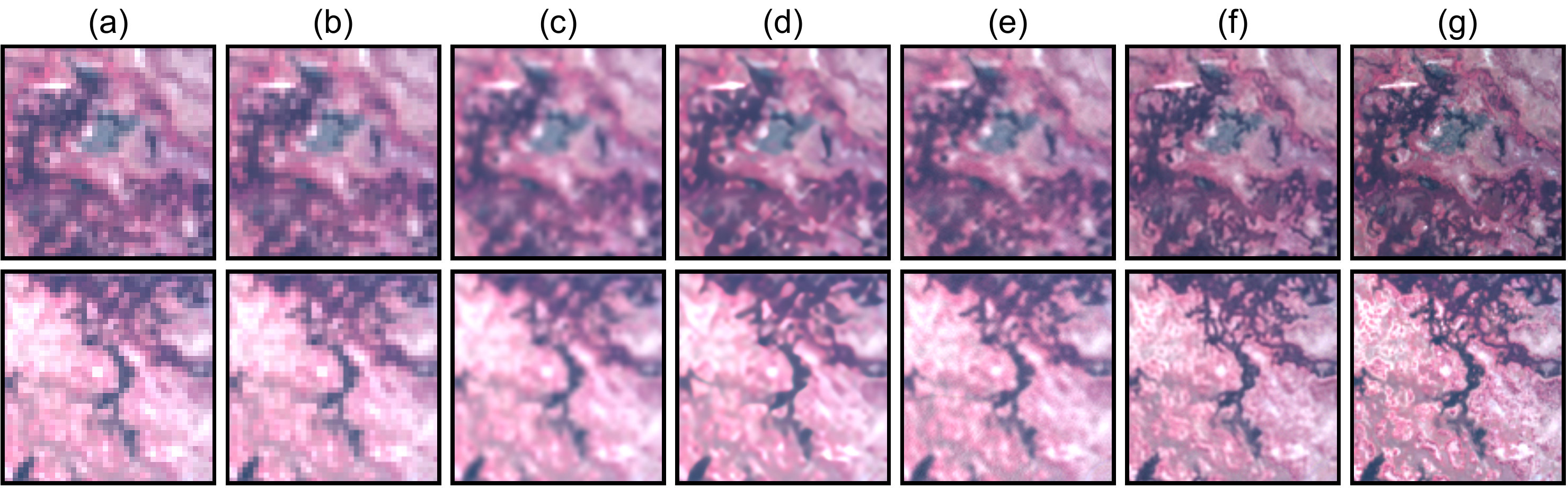}
        \caption{Up-sampled images of 12-th (in first row) and 14-th (in second row)  patch of Botswana dataset. (a) LR-HSI. (b) Nearest-neighbor. (c) Bicubic. (d) LapSRN \cite{LapSRN}. (e) DIP with only spectral energy \cite{DHP-DARN}. (f) DIP with our spatial+spectral energy $(Q_{ss}; \lambda=0.8$). (g) Reference. The RGB image is generated by utilizing the 10-th, 35-th, and 61-th bands of the HSI for blue, green and red bands, respectively.}
        \label{fig:botswana_upsampling_comp}
    \end{figure}
    
    \begin{figure}[tb]
        \centering
        \includegraphics[width=\linewidth]{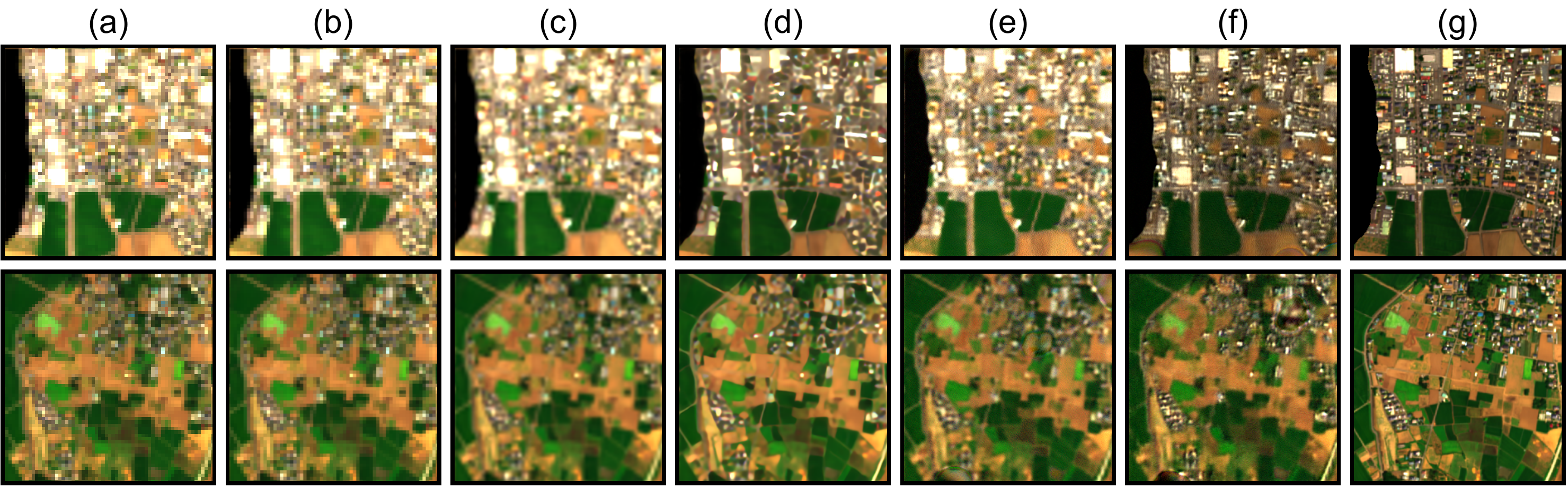}
        \caption{Up-sampled images of 37-th (in first row) and 50-th (in second row)  patch of Chikusei dataset. (a) LR-HSI. (b) Nearest-neighbor. (c) Bicubic. (d) LapSRN \cite{LapSRN}. (e) DIP with only spectral energy \cite{DHP-DARN}. (f) DIP with our spatial+spectral energy $(Q_{ss}$ ($\lambda=0.8$). (g) Reference. The  RGB image is generated by utilizing the 12-th, 20-th, and 29-th bands for blue, green and red bands, respectively.}
        \label{fig:chikusei_upsample_comp}
    \end{figure}
    
    \paragraph*{Discussion on the regularization constant $\lambda$} Let us first consider the case where the regularization constant $\lambda$ is set to zero. This is equivalent to the case where we only have the spectral constraint. In this case, the DIP network minimizes the distance between the down-sampled version of the up-sampled HSI and the LR-HSI. Since the down-sampling operator acts as a low-pass filter in the frequency domain, what DIP network actually minimizes is that the distance between the low-pass version of the up-sampled HSI and the LR-HSI. Because of this reason, the up-sampled HSI from the DIP network trained only with spectral constraint lacks the high frequency components such as edge information and fine structures. Now let us consider the case where we have both spatial and spectral constraint in the DIP loss function. As we described in Section \ref{DIP}, we combined the spatial and spectral constraints via regularization parameter $\lambda$. The value of $\lambda$ controls the fidelity of the predicted PAN image towards the actual PAN image. Since the predicted PAN image and the up-sampled HSI are coupled via spectral response function, to make the predicted PAN image close as possible to the actual PAN image, the DIP network tries to predict some of the high-frequency components such as edges and fine structures in the PAN image,  while maintaining the low-pass version of the up-sampled HSI close to the LR-HSI. Therefore, the regularization constant what actually controls is the amount of high-frequency components fused from PAN image to the up-sampled HSI. This explain the observation that we made from Figure \ref{fig:pavia_lambda_tune_ql}, Figure \ref{fig:botswana_lambda_tune_ql}, and Figure \ref{fig:chikusei_lambda_tune_ql}, where when the value of the regularization parameter increases the DIP network embed some of the high-frequency information to the up-sampled HSI, which ultimately helps to improve the quality of the up-sampled image. However, when the value of the regularization constant is large, the spatial loss term starts to dominate  the loss function, and resulting in drop of spectral-domain performance metrics such as SAM and ERGAS. Therefore, we can achieve high-quality up-sampled HSIs by appropriately controlling the regularization parameter in spatial+spectral energy function.
    
    \subsubsection{Comparison of DIP with the proposed spatial+spectral loss with state-of-the-art up-sampling techniques} In the previous section, we determined the optimal value of the regularization constant $ \lambda $ for our proposed spatial+spectral loss function for the three datasets. In this section, we compare DIP with our spatial+spectral loss against DIP with only spectral loss, and other commonly used up-sampling techniques such as nearest neighbor, bicubic, and LapSRN, both qualitatively and quantitatively. 
    
    Table \ref{tab:pavia_lambda_tune_qt} summarizes the quantitative results of nearest-neighbor, bicubic, LapSRN, and DIP up-sampling methods for the Pavia Center dataset. For this dataset, our proposed DIP method improves the quality of up-sampled images in terms of CC, SAM, RSNR, ERGAS, and PSNR performance measures by $6.6\%$, $37.3\%$, $63.4\%$ $37.3\%$, and $19.5\%$, respectively. We have also noticed that this improvement is accompanied by a drop in the SAM index which is around 1.8\% compared to the DIP with spectral loss. This fall in the SAM index is not that significant compared to the improvements we have achieved in terms of all other performance measures. Further, we can cross-verify these quantitative results with qualitative results that we have shown in Figure \ref{fig:pavia_upsampling_comp} for the Pavia Center dataset. We can see that the DIP up-sampled images with our proposed spatial+spectral constraint looks much more closer to the reference image, and have predicted very fine structures and edges compared to other upsampling methods. 
    
    We also summarize the quantitative results for different up-sampling methods for the Botswana dataset in Table \ref{tab:botswana_lambda_tune_qt}. As we can see, DIP with the proposed spatial+spectral loss improves the quality of up-sampled images in terms of all the performance metrics by a significant margin: CC value increased by 0.4\%, SAM value reduced by 4.3\%, RMSE value reduced by 14.0\%, RSNR value improved by 11.2\%, ERGAS value reduced by 1.7\%, and PSNR value value increased by 5.2\%. Also, we can verify these quantitative results with the qualitative results shown in Figure \ref{fig:botswana_upsampling_comp} for the Botswana dataset. Similar to the qualitative results that we have observed for the Pavia Center dataset, we can see the the up-sampled images using DIP with our proposed spatial+spectral loss is much more closer to the reference HSI. 
    
    Finally, we summarize the quantitative results for different upsampling methods for the Chikusei dataset in Table \ref{tab:chikusei_lambda_tune_qt}. In this case also, the performance of DIP up-sampled images with our proposed spatial+spectral loss outperforms five out of six performance measures that we considered for the analysis. As we can see from the Table  \ref{tab:chikusei_lambda_tune_qt}, our DIP method has increased the value of CC by 0.1\%, has decreased the value of RMSE by 26.3\%, has increased the RSNR by 22.9\%, has decreased the ERGAS by 26.2\%, and has increased the PSNR by 8.9\% over the state-of-the-art results. Similar to the Pavia Center dataset, in this dataset also we have observed that the drop in SAM index; however this is negligible compared to the performance gained in terms of the other quantitative measures. Furthermore, following the similar trend with other datasets, we have included the qualitative results in Figure \ref{fig:chikusei_upsample_comp} for the Chikusei dataset. From the qualitative results also we can see that  DIP with our spatial+spectral constraint is able to predict very fine structures and edges more accurately than the other methods. 
    
    In summary, we have shown that  the DIP method with our proposed spatial+spectral constraints outperforms the state-of-the-art up-sampling methods with a significant margin in all the datasets that we have considered in this study. In the next section, we present final fusion results and compare them with state-of-the-art pansharpening algorithms, qualitatively and quantitatively.
    
    \begin{table}[tb]
        \centering
        \caption{The average quantitative results on the Pavia Center dataset.}
        \begin{tabular}{lp{0.5cm}p{0.5cm}p{0.6cm}p{0.5cm}p{0.6cm}c}
        \hline
        \multirow{3}{*}{Method}                          & CC                & SAM               & RMSE                  & RSNR              & ERGAS             & PSNR  \\
                                        &                   &                   &  $\times 10^{-2}$     &                   &                   & \\
                                        & ($\uparrow$)      & ($\downarrow$)    &  ($\downarrow$)       & ($\uparrow$)      & ($\downarrow$)    & ($\uparrow$)\\
        \hline
        PCA\cite{PCA1}                  & 0.845            & 8.92            & 3.45            & 34.32           & 6.64            & 31.26\\
        GFPCA \cite{GFPCA}              & 0.902            & 8.31            & 3.98            & 29.34           & 7.44            & 29.09\\
        BF \cite{BF}                    & 0.918            & 9.60            & 3.44            & 31.99           & 6.63            & 30.22\\
        BFS \cite{BFS}                  & 0.925            & 8.10            & 3.05            & 34.37           & 6.00            & 31.09\\
        SFIM \cite{SFIM}                & 0.946            & 6.76            & 2.55            & 37.47           & 5.43            & 32.61\\
        GS\cite{GS}                     & 0.961            & 6.62            & 2.55            & 38.08           & 4.95            & 32.93\\
        GSA\cite{GS}                    & 0.950            & 7.15            & 2.34            & 39.60           & 4.70            & 33.52\\
        MTF-GLP-HPM \cite{MTF-GLP-HPM}  & 0.955            & 6.81            & 2.25            & 40.70           & 4.77            & 33.97\\
        CNMF \cite{CNMF}                & 0.960            & 6.64            & 2.20            & 40.79           & 4.39            & 34.14\\
        MTF-GLP \cite{MTF-GLP}          & 0.956            & 6.55            & 2.20            & 40.70           & 4.45           & 34.12\\
        HySure \cite{hysure}            & 0.966            & 6.13            & 1.80            & 44.60           & 3.77            & 35.91\\
        HyperPNN \cite{Hyper-PNN}       & 0.967            & 6.09            & 1.67            & 48.62           & 3.82            & 36.70\\
        DHP-DARN \cite{DHP-DARN}        & 0.969            & 6.43            & 1.56            & 49.17           & 3.95            & 37.30\\
        DIP-HyperKite (ours)            & \textbf{0.980}   & \textbf{5.61}   & \textbf{1.29}   & \textbf{51.72}  & \textbf{2.85}   & \textbf{38.65}\\
        \hline
        \end{tabular}
        \label{tab:pavia_final_fuse_quant}
    \end{table}
    
    \begin{table}[tb]

        \centering
        \caption{The average quantitative results on the Botswana dataset.}
        \begin{tabular}{lp{0.5cm}p{0.5cm}p{0.6cm}p{0.5cm}p{0.6cm}c}
        \hline
        \multirow{3}{*}{Method}                          & CC                & SAM               & RMSE                  & RSNR              & ERGAS             & PSNR  \\
                                        &                   &                   &  $\times 10^{-2}$     &                   &                   & \\
                                        & ($\uparrow$)      & ($\downarrow$)    &  ($\downarrow$)       & ($\uparrow$)      & ($\downarrow$)    & ($\uparrow$)\\
        \hline
        PCA\cite{PCA1}                  & 0.946            & 2.22            & 1.74            & 57.10           & 2.89            & 28.17\\
        GFPCA \cite{GFPCA}              & 0.925            & 2.48            & 1.97            & 53.81           & 3.18            & 26.75\\
        BF \cite{BF}                    & 0.919            & 2.41            & 1.86            & 55.43           & 3.37            & 26.88\\
        BFS \cite{BFS}                  & 0.918            & 2.39            & 1.85            & 55.52           & 3.38            & 26.91\\
        SFIM \cite{SFIM}                & 0.890            & 3.31            & 2.56            & 48.30           & 2.98            & 27.27\\
        GS\cite{GS}                     & 0.949            & 2.17            & 1.68            & 57.55           & 2.74            & 28.32\\
        GSA\cite{GS}                    & 0.964            & 1.86            & 1.28            & 63.02           & 2.16            & 30.78\\
        MGH \cite{MTF-GLP-HPM}          & 0.962            & 1.90            & 1.33            & 62.23           & 2.15            & 30.47\\
        CNMF \cite{CNMF}                & 0.951            & 2.28            & 1.38            & 60.90           & 2.48            & 29.63\\
        MG \cite{MTF-GLP}               & 0.963            & 1.88            & 1.32            & 62.23           & 2.16            & 30.45\\
        HySure \cite{hysure}            & 0.963            & 1.93            & 1.19            & 63.80           & 2.12            & 30.97\\
        HyperPNN \cite{Hyper-PNN}       & 0.957            & 1.92            & 1.06            & 66.22           & 2.40            & 29.00\\
        DHP-DARN \cite{DHP-DARN}        & 0.954            & 1.91            & 1.05            & 66.22           & 2.35            & 29.98\\
        DIP-HyperKite (ours)            & \textbf{0.974}   & \textbf{1.68}   & \textbf{0.96}   & \textbf{67.98}  & \textbf{1.89}   & \textbf{32.12}\\
        \hline
        \end{tabular}
        \label{tab:botswana_final_fuse_quant}
    \end{table}
    
    \begin{table}[tb]
        \centering
        \caption{The average quantitative results on the Chikusei dataset.}
        \begin{tabular}{lp{0.5cm}p{0.5cm}p{0.6cm}p{0.5cm}p{0.6cm}c}
        \hline
        \multirow{3}{*}{Method}         & CC                & SAM               & RMSE                  & RSNR              & ERGAS             & PSNR  \\
                                        &                   &                   &  $\times 10^{-2}$     &                   &                   & \\
                                        & ($\uparrow$)      & ($\downarrow$)    &  ($\downarrow$)       & ($\uparrow$)      & ($\downarrow$)    & ($\uparrow$)\\
        \hline
        PCA\cite{PCA1}                  & 0.297            & 9.99            & 4.47            & 17.86           & 16.70            & 30.96\\
        GFPCA \cite{GFPCA}              & 0.883            & 4.76            & 1.98            & 34.22           & 7.00            & 37.05\\
        BF \cite{BF}                    & 0.903            & 5.15            & 1.94            & 34.40           & 6.62            & 37.89\\
        BFS \cite{BFS}                  & 0.917            & 4.69            & 1.72            & 36.84           & 6.39            & 37.99\\
        SFIM \cite{SFIM}                & 0.928            & 3.79            & 1.43            & 40.51           & 6.43            & 39.55\\
        GS\cite{GS}                     & 0.733            & 5.64            & 2.96            & 26.37           & 8.17            & 35.13\\
        GSA\cite{GS}                    & 0.943            & 3.52            & 1.42            & 40.73           & 4.30            & 41.38\\
        MTF-GLP-HPM \cite{MTF-GLP-HPM}  & 0.929            & 3.82            & 1.45            & 39.38           & 6.40            & 39.85\\
        CNMF \cite{CNMF}                & 0.900            & 4.72            & 1.91            & 36.73           & 5.75            & 39.65\\
        MTF-GLP \cite{MTF-GLP}          & 0.938            & 3.81            & 1.52            & 39.38           & 4.41            & 41.05\\
        HySure \cite{hysure}            & 0.960            & 2.98            & 1.13            & 45.24           & 3.69            & 43.14\\
        HyperPNN \cite{Hyper-PNN}       & 0.946             & 3.97          & 1.11              & 46.55         & 4.77              & 41.57\\
        DHP-DARN \cite{DHP-DARN}        & 0.953            & 3.60            & 1.05            & 46.66           & 4.44            & 42.24\\
        DIP-HyperKite (ours)            & \textbf{0.974}  & \textbf{2.85}   & \textbf{1.03}     & \textbf{46.97}           & \textbf{3.62}            & \textbf{43.53}\\
        \hline 
        \end{tabular}
        \label{tab:chikusei_final_fuse_quant}
    \end{table}
    
    \begin{figure}[tb]
        \centering
        \includegraphics[width=\linewidth]{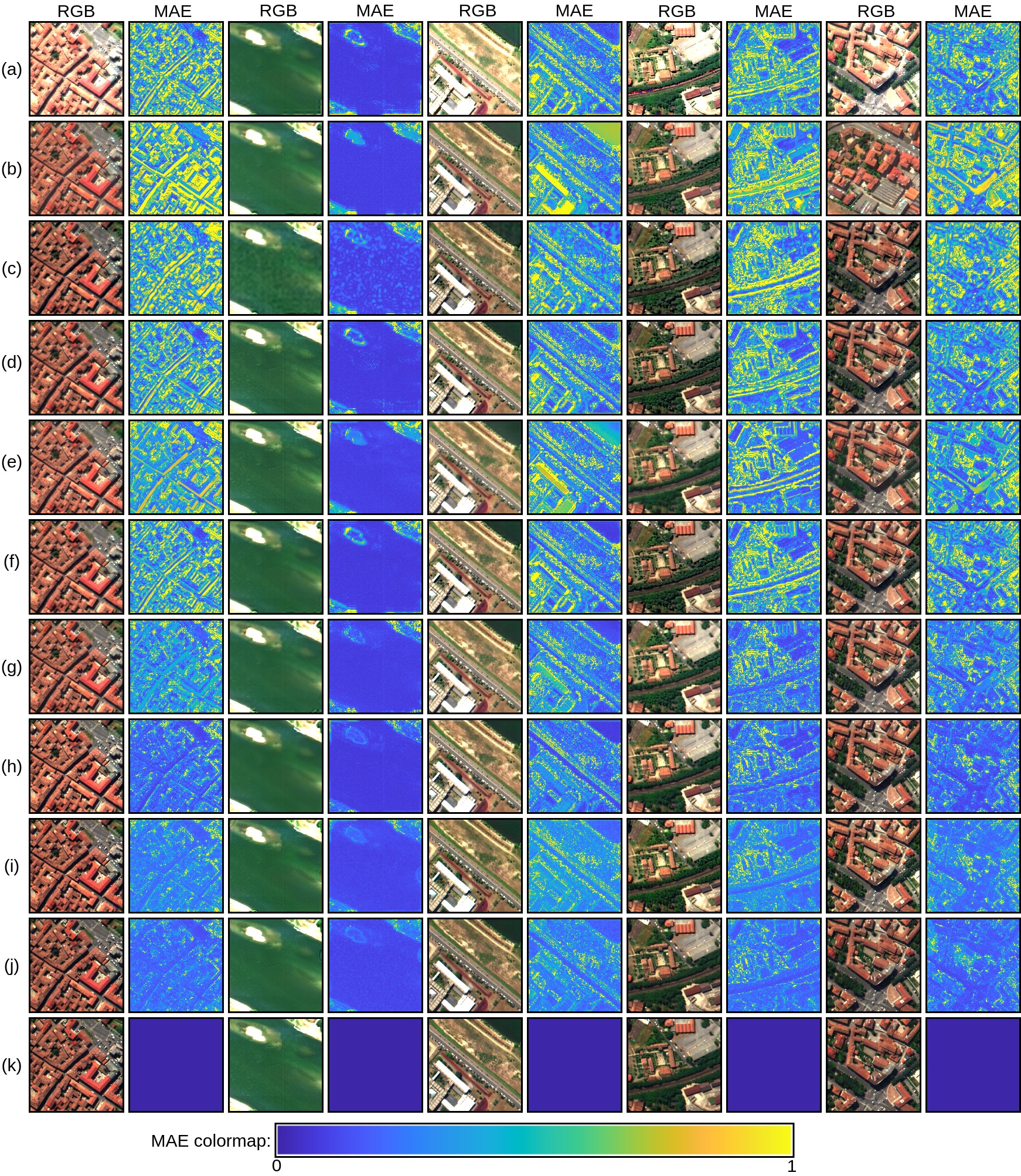}
        \caption{Visual results generated by different pansharpening algorithms for the first (in first and second column), third (in third and fourth column), 12-th (in fifth and sixth column), 20-th (in seventh and eight column), 21-st (in nine and tenth column) patches of the Pavia Center dataset. (a) SFIM \cite{SFIM}. (b) GS \cite{GS}. (c) GSA \cite{GS}. (d) MTF-GLP-HPM \cite{MTF-GLP-HPM}. (e) CNMF \cite{CNMF}. (f) MTF-GLP \cite{MTF-GLP}. (g) HySure \cite{hysure}. (h) HyperPNN \cite{Hyper-PNN}. (i) DHP-DARN \cite{DHP-DARN}. (j) DIP-HyperKite (ours). (k) Reference.}
        \label{fig:pavia_final}
    \end{figure}
    
    \begin{figure}[tb]
        \centering
        \includegraphics[width=\linewidth]{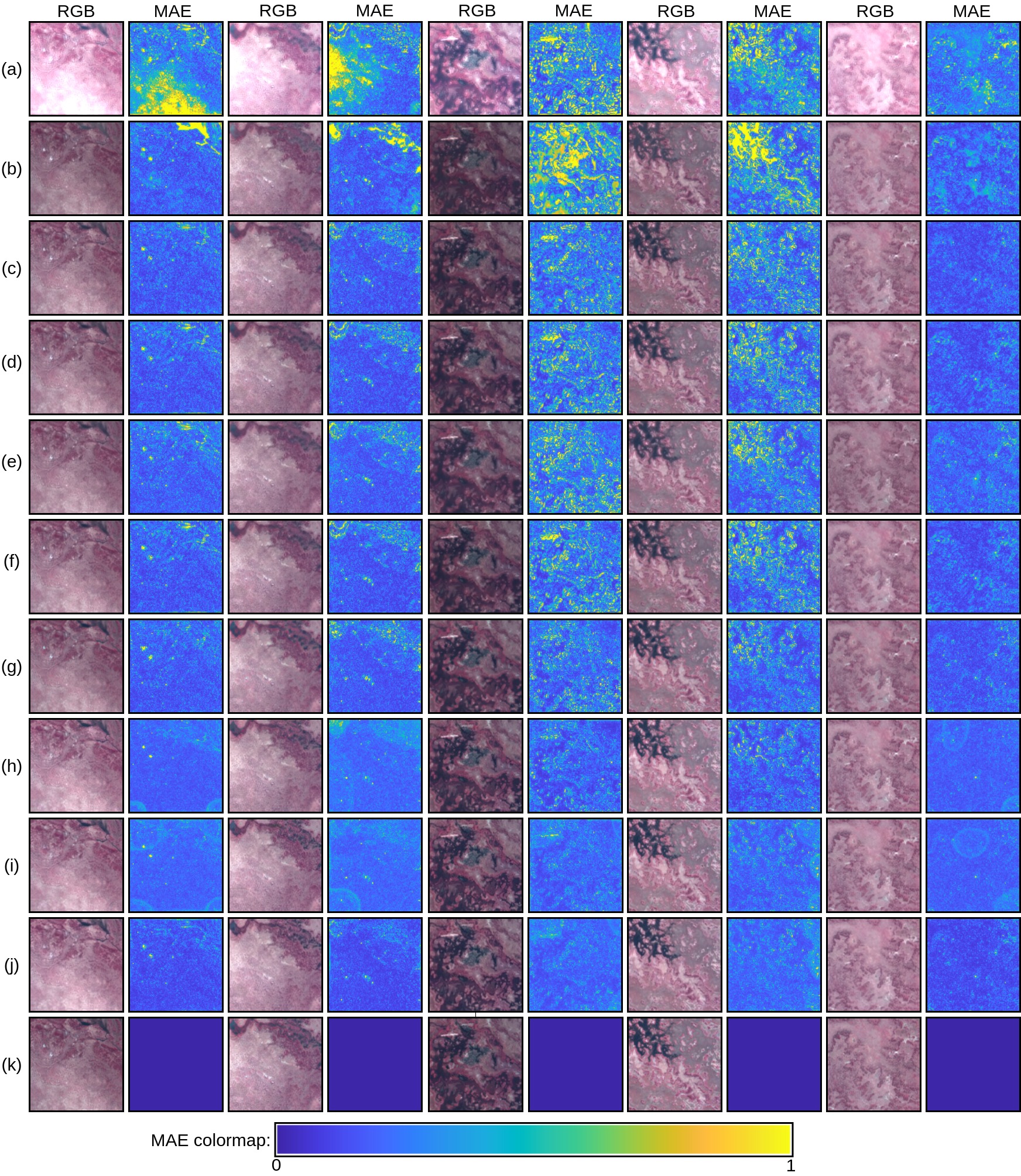}
        \caption{Visual results generated by different pansharpening algorithms for the first (in first and second column), fourth (in third and fourth column), 12-th (in fifth and sixth column), 16-th (in seventh and eight column), 19-st (in nine and tenth column) patches of the Botswana dataset. (a) SFIM \cite{SFIM}. (b) GS \cite{GS}. (c) GSA \cite{GS}. (d) MTF-GLP-HPM \cite{MTF-GLP-HPM}. (e) CNMF \cite{CNMF}. (f) MTF-GLP \cite{MTF-GLP}. (g) HySure \cite{hysure}. (h) HyperPNN \cite{Hyper-PNN}. (i) DHP-DARN \cite{DHP-DARN}. (j) DIP-HyperKite (ours). (k) Reference.}
        \label{fig:botswana_final}
    \end{figure}
    
    \begin{figure}[tb]
        \centering
        \includegraphics[width=\linewidth]{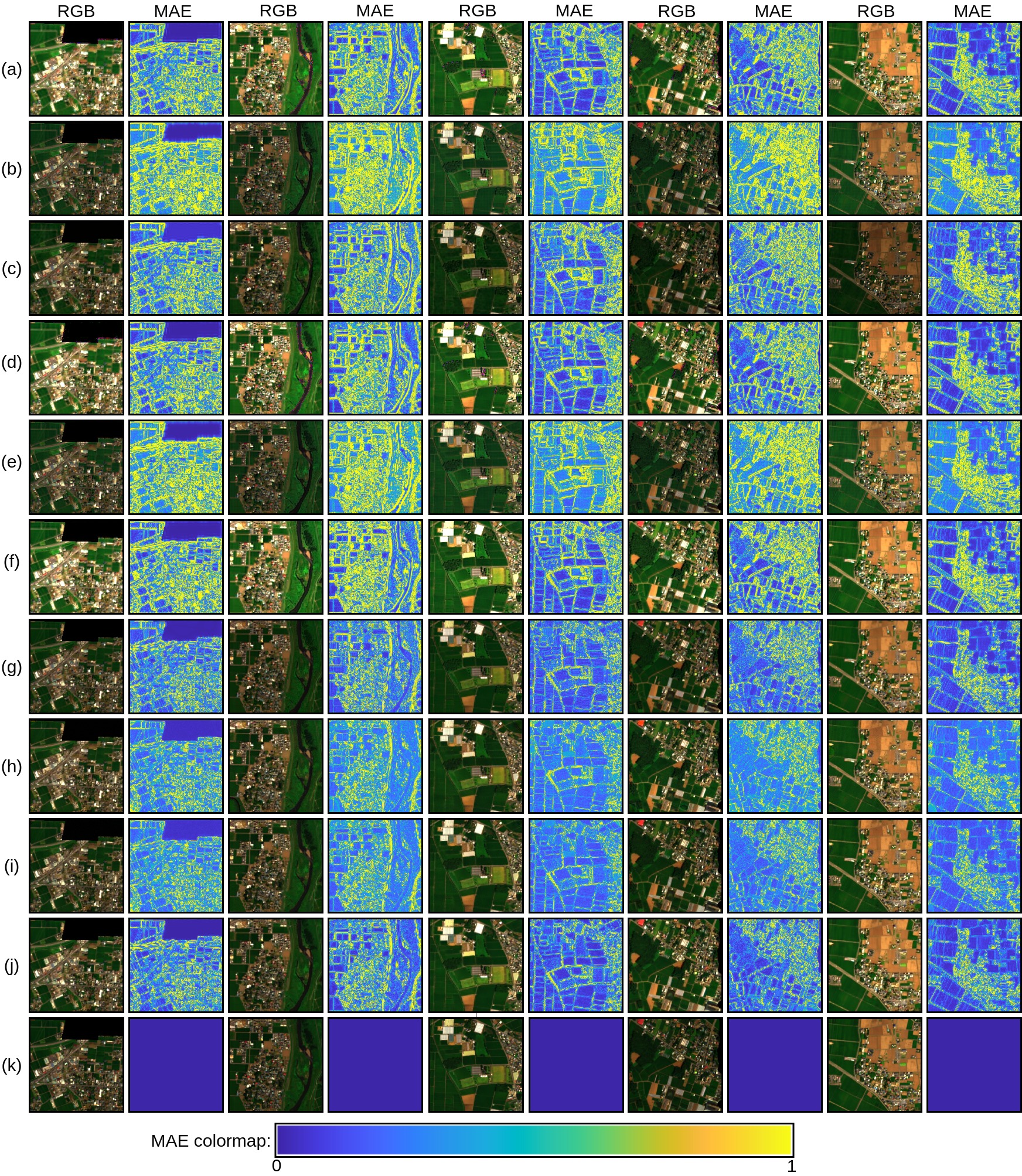}
        \caption{Visual results generated by different pansharpening algorithms for the fifth (in first and second column), 13-th (in third and fourth column), 16-th (in fifth and sixth column), 27-th (in seventh and eight column), 32-nd (in nine and tenth column) patches of the Chikusei dataset. (a) SFIM \cite{SFIM}. (b) BF \cite{BF}. (c) GSA \cite{GS}. (d) MTF-GLP-HPM \cite{MTF-GLP-HPM}. (e) BFS \cite{BFS}. (f) MTF-GLP \cite{MTF-GLP}. (g) HySure \cite{hysure}. (h) HyperPNN \cite{Hyper-PNN}. (i) DHP-DARN \cite{DHP-DARN}. (j) DIP-HyperKite (ours). (k) Reference.}
        \label{fig:chikusei_final}
    \end{figure}
    
\subsection{Final fusion results:}
    \label{sec:final_fusion_results}
    In this section we compare final fusion results from our DIP-HyperKite with the state-of-the-art pansharpening approaches such as PCA \cite{PCA1}, GFPCA \cite{GFPCA}, BF \cite{BF}, BFS \cite{BFS}, SFIM \cite{SFIM}, GS \cite{GS}, GSA \cite{GS}, MTF-GLP-HPM \cite{MTF-GLP-HPM}, CNMF \cite{CNMF}, MTF-GLP \cite{MTF-GLP}, HySure \cite{hysure}, HyperPNN \cite{Hyper-PNN}, and DHP-DARN \cite{DHP-DARN} for Pavia Center, Botswana, and Chikusei datasets.
    
    \paragraph{Final Fusion results on the Pavia Center dataset}
    The average quantitative results for different pansharpening approaches on the testing set of the Pavia Center dataset are shown in Table \ref{tab:pavia_final_fuse_quant}. As can be seen from Table  \ref{tab:pavia_final_fuse_quant}, our proposed HyperKite achieves the highest CC value compared to all the other pansharpening approaches that we have considered in this study. A higher CC value indicates that the fused HSI is closer to the actual HSI with less geometric distortion. Furthermore, our proposed DIP-HyperKite achieved the smallest values for SAM, RMSE, and ERGAS performance measures, indicating the best fusion performance over the other pansharpening approaches. Especially the smallest SAM and ERGAS indicate that our DIP-HyperKite can fuse HSIs with less spectral distortion than the state-of-the-art methods. In addition, our DIP-HyperKite improved the PSNR metric by 3.6\% over the state-of-the-art value. To further verify the fusion quality of our proposed DIP-HyperKite, we present qualitative results in Figure \ref{fig:pavia_final} for the Pavia Center dataset. To better highlight the fusion quality between different pansharpening approaches, we have shown the error plots along with the RGB composite image for each fused HSI. According to the figure, the error maps corresponding to our DIP-HyperKite are much purple than the other pansharpening approaches, indicating minor fusion error. This is mainly because of the ability of our HyperKite network to predict very fine structures and edges by constraining the receptive field of the deep network.
    
    \paragraph{Final Fusion results on the Botswana dataset} 
    The Table \ref{tab:botswana_final_fuse_quant} summarizes the average quantitative results of different fusion methods on the Botswana dataset. Similar to the Pavia Center dataset, we can see that our DIP-HyperKite outperforms all the other HS pansharpening approaches by a considerable margin. Concretely, our DIP-HyperKite has improved the CC by 1.03\%, and PSNR by 3.71\%. In addition, our method has reduced the SAM by 9.68\%, RMSE by 8.57\%, and ERGAS by 10.85\%. Furthermore, we have shown qualitative results related to different pansharpening approaches on Botswana dataset in Figure \ref{fig:botswana_final}. By observing the RGB images and error plots in Figure \ref{fig:botswana_final}, we can see that the fusion results related to our method are much closer to reference image than the other pansharpening approaches.
    
    \paragraph{Final Fusion results on the Chikusei dataset} 
    In this section, we compare the qualitative and quantitative results on the Chikusei dataset. The average quantitative results of different pansharpening approaches on the Chikusei dataset is listed in Table \ref{tab:chikusei_final_fuse_quant}. Similar to the results we have observed for the other two datasets, for this dataset also our proposed DIP-HyperKite outperforms all the pansharpening approaches that we considered for the analysis. Our pansharpening  method improves the CC, SAM, RMSE, RSNR, ERGAS, and PSNR performance measures over the state-of-the-art results by 1.45\%, 19.0\%, 6.67\%, 0.67\%, 18.5\%, and 0.90\%, respectively. To further highlight the fusion quality of our method we present the qualitative results of selected panshaprpening approaches for the Chikusei dataset is shown in Figure \ref{fig:chikusei_final}. By observing the RGB composite image and the error maps we can clearly see that the fusion quality of the proposed DIP-HyperKite is higher than the other pansharpening approaches.

\section{Conclusion}
\label{sec: conclusion}
    In this paper, we have presented a novel approach for HS pansharpening, which mainly consists of three steps: (1) Up-sampling the LR-HSI via DIP, (2) Predicting the residual image via over-complete HyperKite, and (3) Obtaining the final fused HSI by summation. The previously proposed DIP methods for HS up-sampling only impose a constraint in the spectral-domain  by utilizing LR-HSI. To better preserve both spatial and spectral information, we first exploited an additional spatial constraint to DIP by utilizing the available PAN image, thereby introduced both spatial and spectral constraints to the DIP. The comprehensive experiments conducted on three HS datasets showed that our proposed spatial+spectral loss function significantly improved the quality of up-sampled HSIs in CC, RMSE, RSNR, SAM, ERGAS, and PSNR performance measures. Next, in the residual prediction task, we have shown that the residual component between  up-sampled HSI and the reference HSI primarily consists of edge information and very fine structures. Motivated by this observation, we proposed a novel over-complete deep-learning network for the residual prediction task. In contrast to the conventional under-complete representations, we have shown that our over-complete network is competent to focus on high-level features such as edges and fine structures by constraining the receptive field of the network. Finally, the fused HSI is obtained by adding the residual HSI and the up-sampled HSI. The comprehensive experiments conducted on three HS datasets demonstrated the superiority of our DIP-HyperKite over the other state-of-the-art results in terms of qualitative and quantitative evaluations.
    
\ifCLASSOPTIONcaptionsoff
  \newpage
\fi



\bibliographystyle{IEEEtran}
\bibliography{references.bib}
%



%

\begin{IEEEbiography}
[{\includegraphics[width=1.0in,height=1.25in,clip,keepaspectratio]{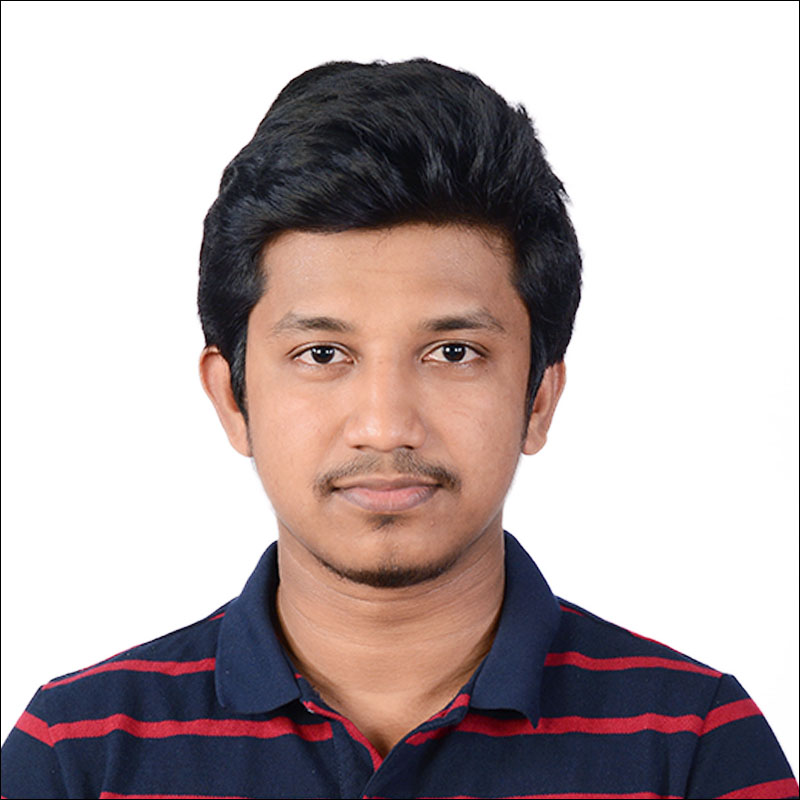}}]%
{Wele Gedara Chaminda Bandara} \text{(Student Member, IEEE)} is a Ph.D. student in the Department of Electrical and Computer Engineering (ECE) at the Johns Hopkins University, USA. Before joining the Johns Hopkins, he worked as a graduate research student in the Department of Electrical and Electronic Engineering at the University of Peradeniya, Sri Lanka, for an NSF-funded project from 2019 to 2020. He graduated from the University of Peradeniya with first-class honors in Electrical and Electronic Engineering in 2019. His research interests include computer vision and image processing with applications in remote sensing, and  hyperspectral image processing.
\end{IEEEbiography}

\begin{IEEEbiography}
	[{\includegraphics[width=1.25in,height=1.25in,clip,keepaspectratio]{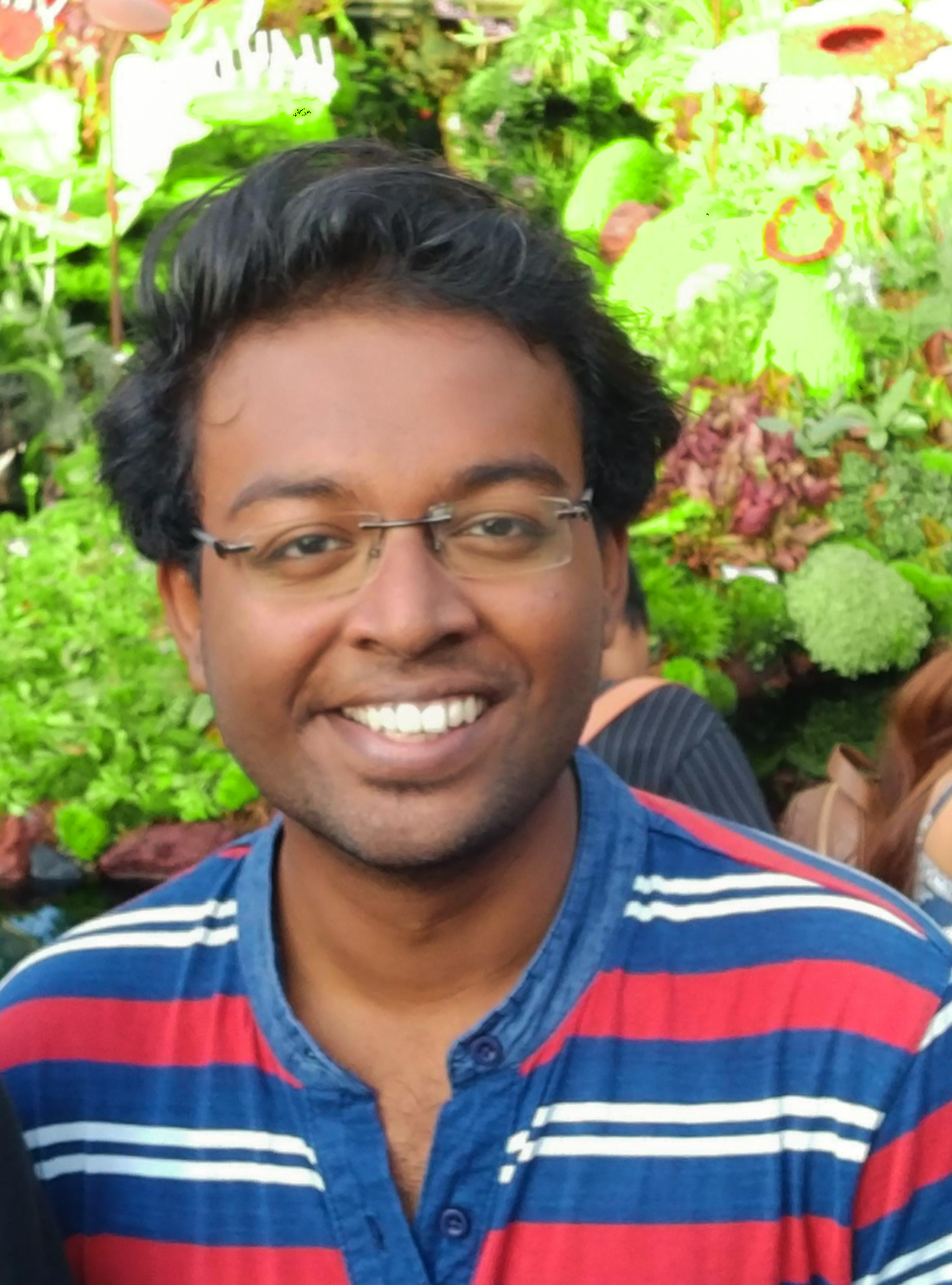}}]%
	{ Jeya Maria Jose Valanarasu}\text{(Student Member, IEEE)} is Ph.D. student in the Department
	of Electrical and Computer Engineering (ECE) at
	Johns Hopkins University, USA. Prior to joining Hopkins, he graduated from NIT Trichy, India in 2019 with a Bachelor’s degree in Instrumentation and Control Engineering. He also spent some time working in the Biomedical Engineering Department at National University of Singapore (NUS) as a visiting research intern.
	His research interests include image/3D segmentation, image enhancement, and image-to-image translation for computer vision and medical imaging tasks.
\end{IEEEbiography}



\begin{IEEEbiography}
[{\includegraphics[width=1.25in,height=1.25in,clip,keepaspectratio]{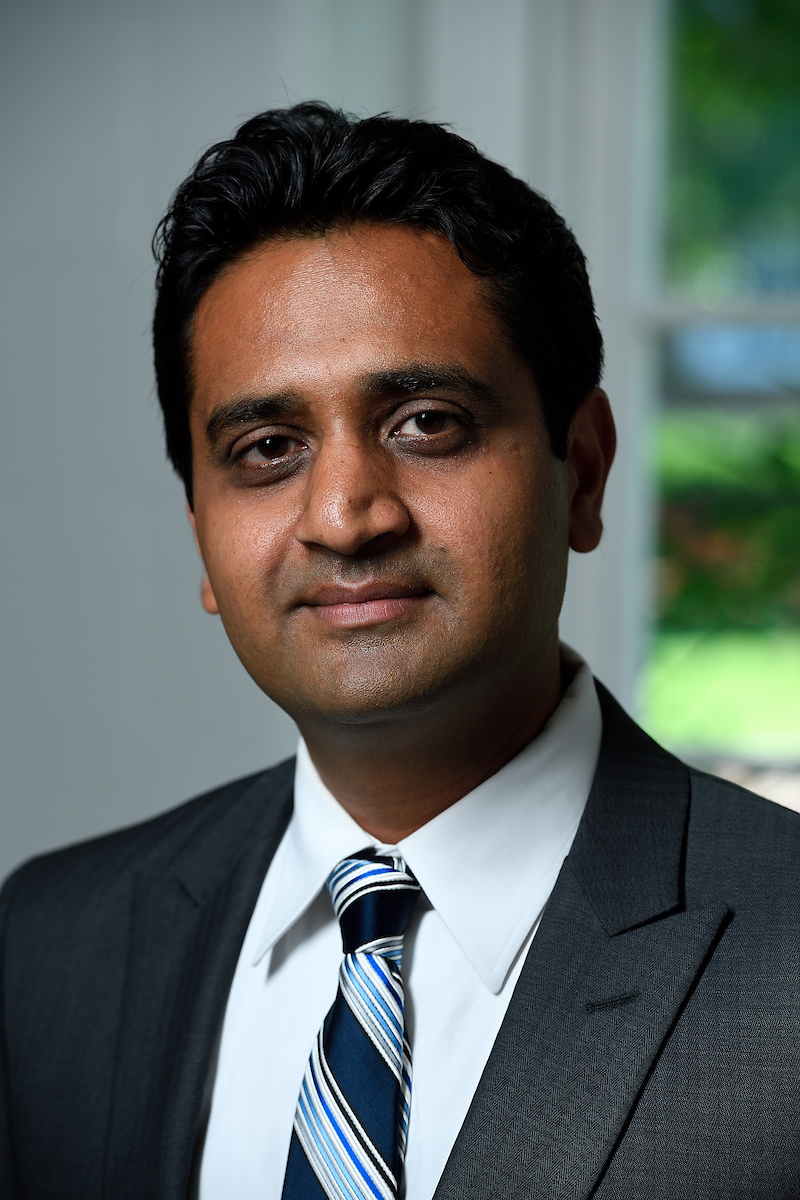}}]%
{Vishal M. Patel}
\text{[SM'15]} is an Associate Professor in the Department of Electrical and Computer Engineering (ECE) at Johns Hopkins University. Prior to joining Hopkins, he was an A. Walter Tyson Assistant Professor in the Department of ECE at Rutgers University and a member of the research faculty at the University of Maryland Institute for Advanced Computer Studies (UMIACS). He completed his Ph.D. in Electrical Engineering from the University of Maryland, College Park, MD, in 2010.  His  current  research interests include signal processing, computer vision, and  pattern  recognition  with  applications  in  biometrics  and  imaging.    He has received a number of awards including the 2021 NSF CAREER Award, the 2016 ONR Young Investigator Award, the 2016 Jimmy Lin Award for Invention, A. Walter Tyson Assistant Professorship Award, Best Paper Award at IEEE AVSS 2017 $\&$ 2019, Best Paper Award at IEEE BTAS 2015, Honorable Mention Paper Award at IAPR ICB 2018, two Best Student Paper Awards at IAPR ICPR 2018, and Best Poster Awards at BTAS 2015 and 2016. He is an Associate Editor of the IEEE Signal Processing Magazine, Pattern Recognition Journal, and serves on the Machine Learning for Signal Processing (MLSP) Committee of the IEEE Signal Processing Society. He serves as the vice president of conferences for the IEEE Biometrics Council. He is a member of Eta Kappa Nu, Pi Mu Epsilon, and Phi Beta Kappa.
\end{IEEEbiography}




\end{document}